\documentclass[lang=english]{tumarxivarticle}

\usepackage{tikz}
\usetikzlibrary{shapes.geometric}
\usetikzlibrary{positioning}
\usetikzlibrary{calc}
\usetikzlibrary{spy}
\usepackage{pgfplots}
\pgfplotsset{compat = 1.3}
\usepackage{pgfplotstable}
\usetikzlibrary{arrows.meta, decorations.markings, math, arrows.meta}
\usepackage{tumcolors}
\definecolor{viridisleft}{HTML}{440154}
\definecolor{viridismiddle}{HTML}{20908D}
\definecolor{viridisright}{HTML}{FDE725}

\usepackage{calc}

\usepackage[font=footnotesize]{subcaption}
\usepackage[
    backend=biber,
    style=numeric,
    maxbibnames=99,
    maxcitenames=1,
    doi=false,
    isbn=false,
    url=false,
    date=year,
    sorting=ydnt,
    giveninits=true
  ]{biblatex}
\bibliography{bibliography.bib}
\newcommand*{\citet}{\textcite}
\newcommand*{\Citet}{\Textcite}
\newcommand*{\citep}{\parencite}

\usepackage{siunitx}
\DeclareSIUnit{\pixel}{px}
\usepackage{amssymb}
\usepackage{amsmath}
\newcommand{\loss}{\ensuremath{\mathcal{L}}}
\newcommand{\lone}{\ensuremath{\ell_1}}
\newcommand{\ltwo}{\ensuremath{\ell_2}}
\newcommand{\loneloss}{\ensuremath{\loss_{\lone}}}

\usepackage{tummath}

\usepackage{csquotes}

\usepackage{etoolbox}
\robustify\bfseries
\usepackage{setspace}

\usepackage{tabularx}
\usepackage{booktabs}
\usepackage{multicol}
\newcommand{\hf}{\scriptsize}
\usepackage{multirow}
\usepackage{colortbl}

\newacronym{sgm}{SGM}{semi-global matching}
\newacronym{gsd}{GSD}{ground sampling distance}
\newacronym{rmse}{RMSE}{root mean squared error}
\newacronym{mse}{MSE}{mean squared error}
\newacronym{miou}{mIOU}{mean intersection over union}
\newacronym{mae}{MAE}{mean absolute error}
\newacronym{dsm}{DSM}{digital surface model}
\newacronym{dem}{DEM}{digital elevation model}
\newacronym{lod2}{LOD2}{level of detail 2}
\newacronym{cnn}{CNN}{convolutional neural network}
\newacronym{lidar}{LiDAR}{light detection and ranging}
\newacronym{uav}{UAV}{unmanned aerial vehicle}
\newacronym{gan}{gan}{generative adversarial network}
\newacronym{cgan}{cGAN}{conditional generative adversarial network}
\newacronym{ilsvrc}{ILSVRC}{ImageNet Large Scale Visual Recognition Challenge}

\usepackage{eso-pic}

\usepackage{cuted}
\usepackage[capitalise]{cleveref}
\crefrangelabelformat{figure}{#3#1#4 to #5\crefstripprefix{#1}{#2}#6}
\crefmultiformat{figure}{figs.~#2#1\xdef\mycreffirstarg{#1}#3}%
 { and~#2{\crefstripprefix{\mycreffirstarg}{#1}}#3}{, #2#1#3}{ and~#2#1#3}

\begin{document}


\AddToShipoutPicture*{%
  \AtTextUpperLeft{%
    \put(0,\LenToUnit{1cm}){%
        \parbox[][][c]{\textwidth}{%
          \centering
          \tikz{
            \node[rectangle, fill=black!10, draw=black!10, thick, minimum width=9.25cm, minimum height=1.75cm]{};
            \node[rectangle, fill=black!10, draw=black!10, thick, minimum width=9.25cm, align=center]{This paper was accepted for publication in the \\ \textbf{ISPRS Journal of Photogrammetry and Remote Sensing}};
          }
      }%
    }%
  }%
}%


\title{
  \normalfont
  A Generalized Multi-Task Learning Approach to \\ Stereo DSM Filtering in Urban Areas
  }

\setkomafont{author}{\large}
\author{
  Lukas Liebel\textsuperscript{\scriptsize 1}\thanks{Corresponding Author: \url{lukas.liebel@tum.de}} \and Ksenia Bittner\textsuperscript{\scriptsize 2} \and Marco K\"orner\textsuperscript{\scriptsize 1}
  }
\publishers{\normalsize
  \textsuperscript{\scriptsize 1}~Computer Vision Research Group,
  Chair of Remote Sensing Technology \\
  Technical University of Munich (TUM), Germany \\[0.25cm]

  \textsuperscript{\scriptsize 2}~Photogrammetry and Image Analysis,
  Remote Sensing Technology Institute \\
  German Aerospace Center (DLR), Germany
}
\date{}

\maketitle

\begin{strip}
  \vspace{-0.75cm}

  \renewcommand{\abstractname}{}
  \begin{onecolabstract}

    City models and height maps of urban areas serve as a valuable data source for numerous applications, such as disaster management or city planning.
    While this information is not globally available, it can be substituted by \glspl{dsm}, automatically produced from inexpensive satellite imagery.
    However, stereo \glspl{dsm} often suffer from noise and blur.
    Furthermore, they are heavily distorted by vegetation, which is of lesser relevance for most applications.
    Such basic models can be filtered by \glspl{cnn}, trained on labels derived from \glspl{dem} and 3D city models, in order to obtain a refined \gls{dsm}.
    We propose a modular multi-task learning concept that consolidates existing approaches into a generalized framework.
    Our encoder-decoder models with shared encoders and multiple task-specific decoders leverage roof type classification as a secondary task and multiple objectives including a conditional adversarial term.
    The contributing single-objective losses are automatically weighted in the final multi-task loss function based on learned uncertainty estimates.
    We evaluated the performance of specific instances of this family of network architectures.
    Our method consistently outperforms the state of the art on common data, both quantitatively and qualitatively, and generalizes well to a new dataset of an independent study area.

  \end{onecolabstract}
  \vspace{1cm}

\end{strip}

\glsresetall


\section{Introduction}
\label{sec:introduction}

\Glspl{dsm} provide insight into urban structures and are, thus, a valuable source of information for authorities, industries, and non-profit organizations alike.
Numerous applications, such as urban planning, disaster response, or environmental simulation, depend on such 3D elevation models at global scale.
However, availability is heavily restricted due to high expenses for acquiring suitable source data through \gls{lidar} measurements or even in-situ surveying, especially in less developed regions.
Photogrammetric stereo reconstruction from \glspl{uav}, aerial, or even high-resolution satellite imagery provides similar results, while merely relying on comparably inexpensive data and automatic processing.
The resulting height maps are, however, affected by noise and inaccuracies due to matching errors, temporal changes, or interpolation.
Most relevant for many applications are precise building shapes which are, unfortunately, often obscured beyond recognition.
Furthermore, vegetation, that is of lesser interest and even obstructive for most applications, heavily influences height maps and often covers or distorts the appearance of adjacent buildings.
Removing noise and vegetation from automatically produced stereo \glspl{dsm} while simultaneously refining building shapes yields elevation maps that can serve as a simple geometric proxy for city models in \gls{lod2}\citep{Kolbe2009}.
Methods from computer vision first allowed for successfully tackling this task by making use of state-of-the-art machine learning techniques such as \glspl{cnn} and \glspl{cgan}, as well as training data generated from \glspl{dem} and CityGML models \citep{Kolbe2009}, available to the public through open data initiatives \citep{bittner2018dsm}.

Harnessing the abundance of data provided by an ever-growing number of high-resolution spaceborne imaging sensors and open data sources allows employing high-capacity models that extract usable information from vast datasets.
At the same time, they are able to deal with large intra-dataset variance, in particular, imperfect labels stemming from temporal and spatial offsets between data from different sources.
Finally, models that have been trained on such datasets and have undergone rigorous and genuine evaluation procedures on independent validation and test data are expected to generalize well to large and various areas.

\begin{figure*}
  \centering
  \includegraphics{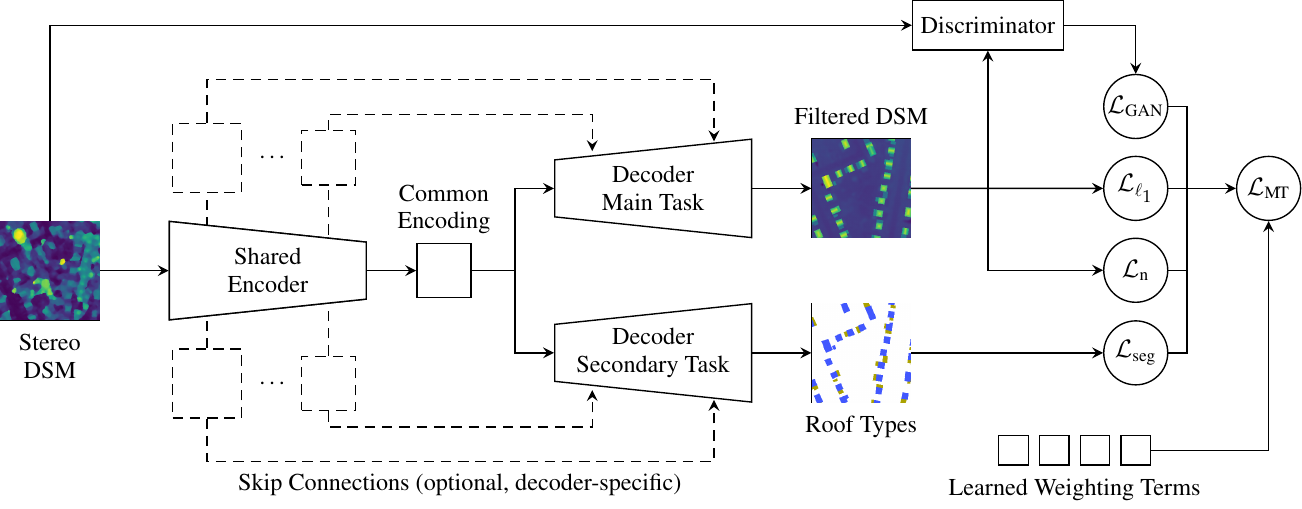}

  \caption{Schematic overview of our generalized approach to multi-task multi-objective DSM filtering and roof type classification.
  The modular encoder-decoder networks consist of a common encoder and task-specific decoder modules.
  Various objectives, evaluating geometrical and semantic properties as well as conditional adversarial feedback, can be utilized and are combined to a loss function with balancing terms that can optionally be learned based on a task uncertainty measure.}
  \label{fig:mt_concept}

\end{figure*}

The results of first approaches in this field were promising but still suffered from improperly removed vegetation and distorted or missing buildings.
These problems can effectively be tackled by multi-task learning approaches that use additional supervision for identifying built-up areas through the secondary task of building footprint segmentation and roof type classification.
This auxiliary task mainly serves as a regularization measure that supports optimization of the main task by restricting the space of possible solutions \citep{Liebel18}.
Since the tasks are inherently related, intuitively, training a joint model for multiple tasks can increase the performance of all tasks.
An approach making use of this technique for \gls{dsm} filtering has recently been proposed \citep{bittner2019multi}.
Experimental results proved the effectiveness of the method, yet there remain some notable issues in the prediction results.
As they still hint at insufficient utilization of the available supervision as a regularization measure, we revise this approach with a more comprehensive abstracted or generalized multi-task concept, illustrated in \cref{fig:mt_concept}.
To this end, we designed a modular encoder-decoder model with a single common encoder and multiple task-specific decoders.
By sharing a major portion of the network parameters, we allow for the optimization of a common representation that is valuable for all tasks.
Multiple objectives per task enable the simultaneous evaluation of different aspects, such as geometrical properties, semantics, or conditional adversarial terms.
Secondary tasks, such as roof type segmentation, mainly serve as a regularization measure to the highly non-convex and hard optimization problem of the main regression task.
All parts of this generic model can be deactivated on demand and the architecture of encoder-decoder blocks can be selected and adjusted individually.
Finally, automatic balancing of the single-objective and single-task terms in the multi-objective multi-task loss function is achieved by optimizing weighting terms along with the network parameters based on an uncertainty measure.
This allows for the utilization of arbitrarily many terms in the loss function without requiring manual balancing of their individual contribution to the final objective, subject to optimization.
As a consequence, our approach scales well to more tasks and objectives easily.
We expect the proposed system to exploit and leverage provided supervision efficiently and effectively.
Prior methods \citep{bittner2018dsm, bittner2018automatic, bittner2019multi} can be interpreted as specific instances of our abstract general framework.
In extensive experiments, we evaluated the suitability of selected instances, which utilize numerous combinations of objectives and decoder architectures, towards improving the predictive performance on unseen data with a focus on mending the remaining systematic problems of previous approaches.

Our main contributions are three-fold.
\begin{enumerate}
  \item We propose a modular multi-task framework consisting of an encoder-decoder model with a shared encoder and task-specific decoders that can optionally be extended by various objectives and tasks.
  Multi-task approaches proposed in prior work can be modeled as instances of this generalized framework.
  \item We introduce the automatic and dynamic balancing of objective terms in the multi-task loss functions by learned weights to the application of \gls{dsm} filtering.
  \item We present novel instances of our framework that are able to undercut the remaining regression error in the state of the art for \gls{dsm} filtering by as much as \SI{34}{\percent}.
\end{enumerate}

The remainder of this paper is structured as follows.
We review the state of the art in both, \gls{dsm} filtering and multi-task learning in the next section.
Our approach to tackling the remaining problems through a modular encoder-decoder system with multiple objectives and learned balancing terms is presented detail in \cref{sec:approach}.
The properties of the datasets used in our experiments are specified in \cref{sec:data} along with the covered study areas and the procedure for deriving pairs of input and ground-truth samples from remote sensing imagery and virtual 3D city models.
In \cref{sec:experiments}, we describe our implementation of the proposed approach with architectural details and derive our final multi-task multi-objective loss function from individual terms and an automated weighting scheme.
We define an experimental setup for training and validating these models and show training results and ablation studies that allow for selection of the most promising models.
These were deployed for the final testing stage, which is described in \cref{sec:evaluation}.
Test results for both study areas are shown, discussed, and compared to the state of the art.
Finally, in \cref{sec:conclusion}, we conclude with a summary of the most important findings and review the suitability of the proposed method for the target application.

\section{Related Work}
\label{sec:stateoftheart}

Multi-task learning \citep{Caruana97} is a machine learning concept that comprises techniques for exploiting mutual information from different supervisory signals, most commonly through neural networks with shared parameters \citep{Ruder17}.
Various machine learning applications, from natural language processing \citep{collobert2008unified} to computer vision, have profited from multi-task approaches.
In the field of computer vision, numerous works explored specific use-cases and combinations of tasks.
A popular example is Fast-RCNN \citep{girshick2015fast} which combines bounding box regression and image classification into a holistic object detection approach.
\Citet{Eigen15} utilize joint learning of single-image depth estimation with surface normal estimation and semantic segmentation.
Similarly, \citet{Kendall18} improved depth estimation results for road scenes by evaluating semantic and instance labels at the same time.
A different approach on single-image depth estimation using multi-task learning has been pursued by \citet{Liebel19} who posed this natural regression task as the classification of discrete depth ranges as an additional auxiliary task and jointly solved for both targets.
Other problems that have recently been tackled by multi-task learning include facial landmark detection \citep{Zhang14} and person attribute classification \citep{Lu17}.
More generally, \citet{taskonomy2018} explored the relationship between visual tasks in general.
Moreover, studies have even been conducted towards the benefit of simultaneously solving seemingly unrelated tasks \citep{Liebel18,Chennupati19}.

While such application-based studies prove the suitability of the multi-task concept for diverse real-world use-cases, recent methodological contributions further advanced the underlying techniques.
Especially active is a branch of research devoted to designing loss functions for multi-task systems, either by developing automated task weighting strategies \citep{Kendall18, Guo18} or by directly training with multiple loss functions through multi-objective optimization \citep{Sener18}.
\Citet{Zhao18} addressed measures for avoiding an unwanted side-effect, often referred to as destructive inference, which occurs when different tasks compete with each other by yielding opposed gradients and thus undermine learning of the whole model.

Urban height estimation from satellite images is a hot research topic in the field of remote sensing, as its availability can positively influence the understanding of urban environments.
Buildings are one of the prominent classes of terrestrial objects within cities.
The appearance or destruction of such is regarded with great interest.
However, estimating the height and exact shape of individual buildings in a city is a time-consuming process that needs a great deal of effort when done manually.
To increase the extraction efficiency of building representations in urban areas, automatic remote sensing methods have been developed.
Yet, estimation from photogrammetric \glspl{dsm}, which are often used for building height and silhouette assessment, also presents some challenges.
Most prominently, noise and mismatches, due to stereo-matching failures in areas with occlusions, low-texture or perspective differences in a stereo image pair, can lead to significant distortions.

Earlier approaches investigated the applicability of filtering techniques \citep{wang1998applying,walker2006comparative,arrell2008spectral} to detect and remove outliers from photogrammetric \glspl{dsm}.
Although they were able to reduce the number of spikes and blunders in the resulting \gls{dsm}, a side-effect was observed in the form of steepness smoothing of building walls.
\Citet{krauss2010enhancement} approached the enhancement of \glspl{dsm} by using segmentation results extracted from stereo images.
Mainly, statistical and spectral information was utilized to approximately detect above-ground objects for their further classification and filtering.
Several methodologies \citep{sirmacek2010enhancing,sirmacek2010detecting} are based on fitting shape models, by either fitting a single rectangular box to a coarse building segment or a chain of such to model more complicated building forms.
Nevertheless, even though the outlines of the extracted buildings became more realistic, the true roof forms were not modeled, as only a single height value was assigned per bounding box, representing a building.

Recent deep learning-based methodology achieves state-of-the-art results in remote sensing image processing, especially by analyzing data with spectral information.
However, the processing or generation of images with continuous values, such as height information in \glspl{dsm}, remains an open problem.
Solutions to this challenge were proposed recently, \ia \citet{ghamisi2018img2dsm} made an attempt to simulate elevation models from single optical images using \gls{cnn} architectures based on \glspl{cgan}.
Although their research is close to our area of interest, the main difference is the employed type of source data.
High-resolution aerial images have been used to estimate height information in their approach.
Since single spectral images do not contain explicit height information, only relative elevation can be estimated based on empirical knowledge.
Using a \gls{dsm} as input, as in our work, only requires learning a residual to yield absolute height values.

In our first work introducing a concept for \gls{dsm} estimation with improved building shapes \citep{bittner2018automatic} we proposed a network based on the \gls{cgan} model by \citet{Isola17}.
A low-quality photogrammetric \gls{dsm} from multi-view stereo satellite imagery serves as input to this network.
Our method is able to reconstruct elevation models with improved building roof forms but artifacts negatively influence the quality of the obtained results.
In a follow-up work \citep{bittner2018dsm}, we improved our method by introducing least-square residuals in the objective function instead of the negative log-likelihood.
This allows reconstructing improved building structures with fewer artifacts and less noise.
To further enhance the assessment of building silhouettes in height images, we recently proposed to add additional supervision through a multi-task learning approach \citep{bittner2019multi}.
Mainly, the simultaneous learning of \gls{dsm} filtering and prediction of roof types masks via an end-to-end learning framework allows to obtain extra information from both tasks and to reconstruct building shapes with more complete structure.
However, the proposed \gls{cgan}-based network architecture was fixed to a specific configuration with the shared part only containing very few layers.
Thus, in fact, the single tasks can not make use of mutual information extracted from the input data, practically leading to two separate networks, independently optimized in parallel.
Besides, the weighting parameters for tuning the contribution of each task in the loss function were experimentally selected and fixed during training.

To compensate for the aforementioned drawbacks, in this work, we propose a more general approach which comprises the properties of prior work while introducing a holistic modular framework.
In our abstracted model, arbitrary combinations of objective functions and tasks can be employed, tasks can be added on-demand, a clear separation into a common part and task-specific parts of the network is defined, various architectures can be used for the respective encoder and decoder blocks of the network, and each contributing objective can optionally be automatically and dynamically weighted by uncertainty-based balancing terms.
Prior methods can be modeled as instances of our generalized approach by employing the respective architectural blocks as encoder and decoder modules, the utilized objectives, and fixed task weights.
Since our framework is much more flexible, it allows for the optimization of instances towards the demands of \gls{dsm} filtering and simultaneous roof type classification without being restricted to this set of tasks or this specific application.

\section{A Modular Multi-Task Framework for DSM Filtering}
\label{sec:approach}

Refining a \gls{dsm} requires the accurate detection of buildings and the ground plane in order to improve valuable features and remove unwanted phenomena, such as noise and vegetation.
Training a deep \gls{cnn} for this task using a simple reconstruction loss is feasible, yet hardly captures all of the desired properties.
Prior work shows that, by adding additional objectives, \eg the direction of surface normals, the quality of predictions can be enhanced \citep{bittner2019multi}.
Adding an auxiliary task to the network complements the multiple objectives for the main task.
The roof type segmentation task, as utilized in prior work, uses ground-truth data derived from the same source as the \gls{dsm} filtering task, thus, adding little overhead to the data creation process.
While the different objectives for the main task evaluate the same network output and therefore consequently optimize the same set of network parameters, the auxiliary task can only share a part of the network weights as an independent output has to be produced.
We approach the design of a suitable network architecture by creating a family of networks through a modular encoder-decoder structure, in which different architectural blocks can be used for each part.
In these encoder-decoder networks, sharing a common encoder enables learning a rich representation while taking into account the supervision of both tasks.
Multiple task-specific decoders allow for appropriate decoding of the shared features for each task.
For our multi-objective multi-task loss function, we balance the contribution of each objective from both tasks by learned weighting factors.
\Cref{fig:mt_concept} illustrates an abstract overview of this abstracted multi-task concept.

\subsection{Multi-Task Concept}

Image-to-image problems, such as \gls{dsm} filtering, require producing an equally-shaped raster output from an input image.
For such tasks, auto-encoders can be employed that map the input to a latent space in a first encoding stage and decode this representation to the final output subsequently.
Since the defined tasks are related by their connection in the physical world, we expect them to rely on similar features in the input image space.
Hence, a major portion of the network parameters can be shared.
By hard-sharing a subset of parameters, each task contributes towards learning an optimal set of generic parameters.
Intuitively, sharing a common encoder enables the extraction of generic features, while individual decoders allow for a task-specific interpretation of the latent representation and translation to the respective outputs.
This auto-encoder-based system with shared and individual parts can be seen in \cref{fig:mt_concept}.

In our implementation, we designed a modular system which is able to make use of various architectures in the encoder and decoder parts of the network.
Hence, we can adapt the network to the different requirements of the main regression and the auxiliary segmentation tasks.
Using encoders of different capacity greatly adds to the variability of our system through changes in the proportion of shared \vs task-specific parameters.

For our experiments, we selected multiple state-of-the-art semantic segmentation architectures as a basis for the decoders.
These highly specialized architectures excel in transforming input images to the desired output map of pixel-wise classification results in close-to-original resolution.
Similar to implementations of numerous semantic segmentation approaches, we use state-of-the-art classification architectures as a backbone for feature extraction in our encoder.
In order to retain the spatial structure of the input, we create a resolution-preserving encoder by modifying the backbone network architecture slightly.
Combinations of different decoder modules yield a vast number of instances of our abstract model.

\subsection{Learning Multiple Objectives}

\Gls{dsm} filtering can be tackled as a single-task regression task, \eg by using an auto-encoder with a reconstruction objective, such as an \lone{} loss \loneloss{}.
In our generalized framework, \cf \cref{fig:mt_concept}, this could be modeled as an encoder followed by a single decoder module and directly optimized using the reconstruction loss.
In order to enhance the quality of the predicted \glspl{dsm}, additional objectives, such as the evaluation of surface normals $\loss_\text{n}$ or adversarial terms $\loss_\text{GAN}$, can be employed \citep{bittner2019multi}.
Such loss functions \loss{} operate independently on the predicted \gls{dsm} and are later combined to a multi-objective loss function.
We implement all three mentioned objectives for the main \gls{dsm} filtering task and integrate them into a single multi-objective multi-task loss $\loss_\text{mt}$ for optimization together with the segmentation objective $\loss_\text{seg}$ of the auxiliary roof type classification task.

\subsection{Learning Task Weights}
\label{subsec:learned_task_weighting}

Since our approach makes use of multiple objectives for both, different objectives and tasks, a strategy for combining them to a loss function for optimization is required.
Balancing the contributing objectives is a vital part of multi-task learning.
Most commonly, this is realized by simply calculating a weighted sum of the single-task losses with fixed balancing factors.
These terms may even be equal in the most simple case.
Since the single objective functions are inherently different in terms of their output range and sensitivity to changes, weighting each of them by an individually adapted balancing factor is an appropriate measure.
While such balancing factors can be tuned along with other hyperparameters to achieve better results, finding a suitable set of weights is a tedious and challenging problem.

This procedure can be replaced by an automated and potentially dynamic weighting scheme.
Multiple solutions have been proposed in the literature, such as weighting based on task difficulty \citep{Guo18}, or task uncertainty \citep{Kendall18}.
We follow the latter that allows for learning task weights $w$ based on an uncertainty measure $\sigma$ along with the network parameters, as indicated in \cref{fig:mt_concept}.
Minimizing a sum of weighted losses $\loss_\text{mt} = \sum_\tau w_\tau \cdot \loss_\tau$ with learnable parameters $w_\tau$ for each task $\tau$ with $\mathcal{T} = \{ \tau_i\}_i$ favors trivial solutions for $w_\tau \rightarrow - \infty$.
Hence, regularization terms $r$ have to be introduced.
As proposed by \citet{Kendall18}, we formulate a multi-task loss
\begin{align}
    \loss_\text{mt} = \sum_\tau \loss_\tau \cdot w_\tau + r_\tau
    \label{eq:kendall_loss}
\end{align}
with $w_\tau = 0.5 \cdot \exp(-\log(\sigma_\tau^2))$ for regression tasks, $w_\tau = \exp(-\log(\sigma_\tau^2))$ for classification tasks and $r_\tau = 0.5 \cdot \log(\sigma_\tau^2)$.

\section{Data Sources and Produced Datasets}
\label{sec:data}

\begin{figure}
  \hfill
    \begin{subfigure}[b]{.35\linewidth}
        \begin{tikzpicture}
            \clip (-.5\linewidth,-.35\linewidth) rectangle (.5\linewidth,.35\linewidth);
            \node at (0,0) (background_image) {\includegraphics[width=\linewidth]{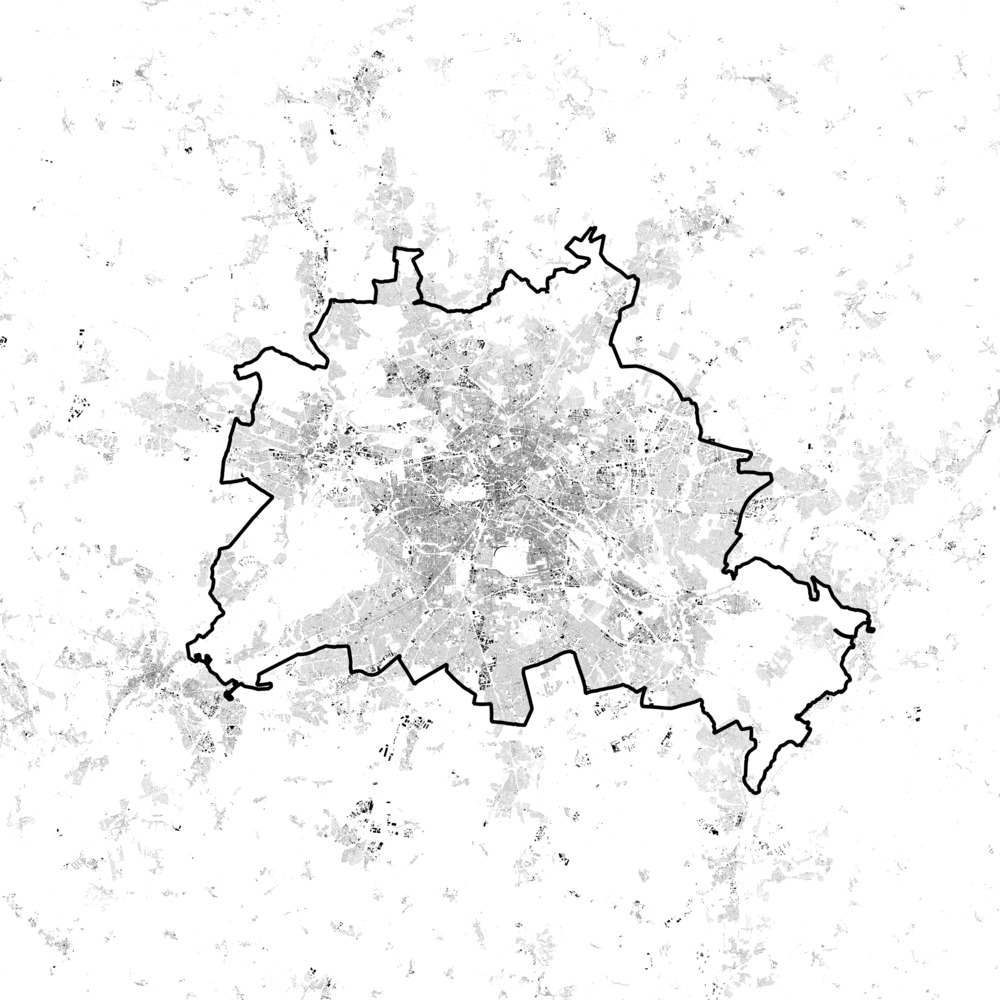}};
            \node at (-0.164112903225806\linewidth, 0.181048387096774\linewidth) (lo) {};
            \node at (-0.164112903225806\linewidth, -0.147983870967742\linewidth) (lu) {};
            \node at (0.120161290322581\linewidth, -0.147983870967742\linewidth) (ru) {};
            \node at (0.120161290322581\linewidth, 0.181048387096774\linewidth) (ro) {};
            \fill[orange, opacity=0.5] (lu) rectangle (ro);
        \end{tikzpicture}
        \caption{Berlin}
        \label{fig:Berlin}
    \end{subfigure}
    \hfill
    \begin{subfigure}[b]{.35\linewidth}
        \begin{tikzpicture}
            \clip (-.5\linewidth,-.35\linewidth) rectangle (.5\linewidth,.35\linewidth);
            \node at (0,0) (background_image) {\includegraphics[width=\linewidth]{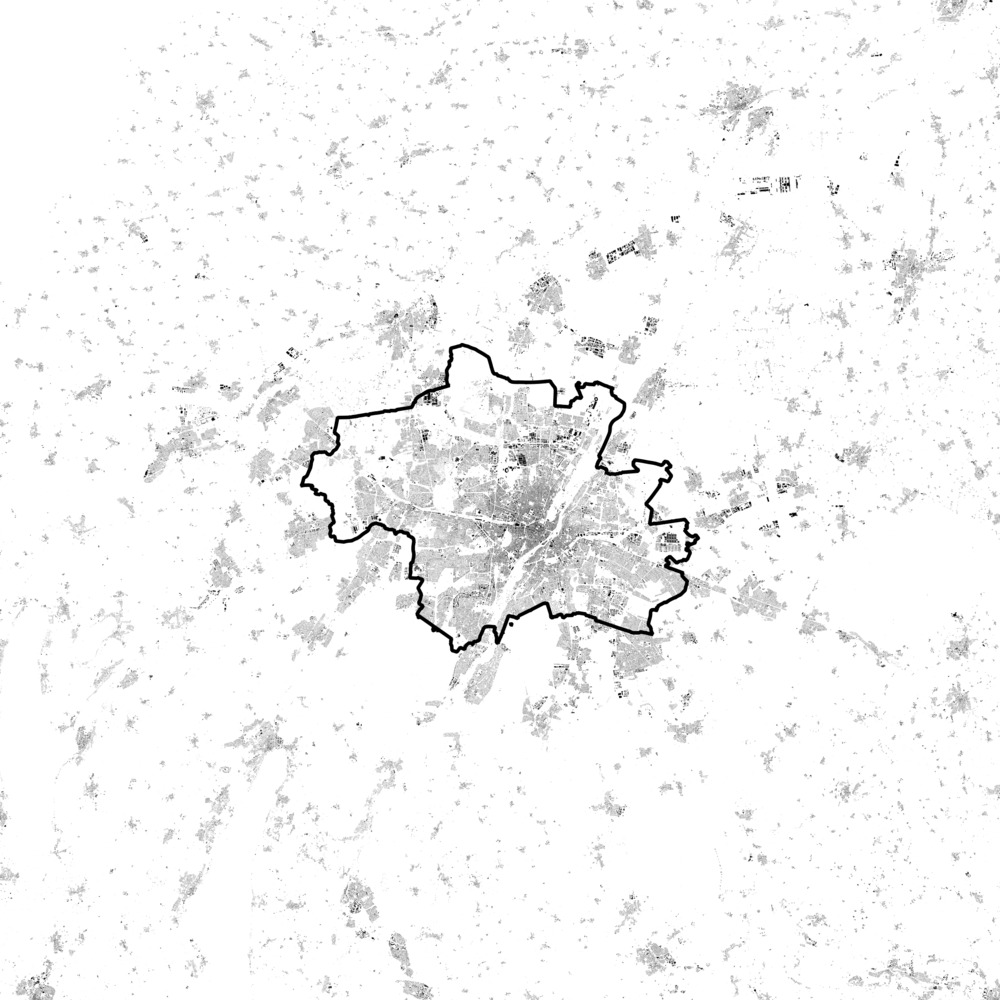}};
            \node at (0.025403225806452\linewidth,0.028629032258065\linewidth) (lu) {};
            \node at (0.02741935483871\linewidth,0.057661290322581\linewidth) (lo) {};
            \node at (0.060483870967742\linewidth,0.054838709677419\linewidth) (ro) {};
            \node at (0.058467741935484\linewidth,0.025806451612903\linewidth) (ru) {};
            \fill[orange, opacity=0.5] (lu.center) -- (lo.center) -- (ro.center) -- (ru.center) -- cycle;
        \end{tikzpicture}%
        \caption{Munich}
        \label{fig:Munich}
    \end{subfigure}
    \hfill
    \begin{subfigure}[b]{.25\linewidth}
      \centering
      \includegraphics[width=.68\linewidth]{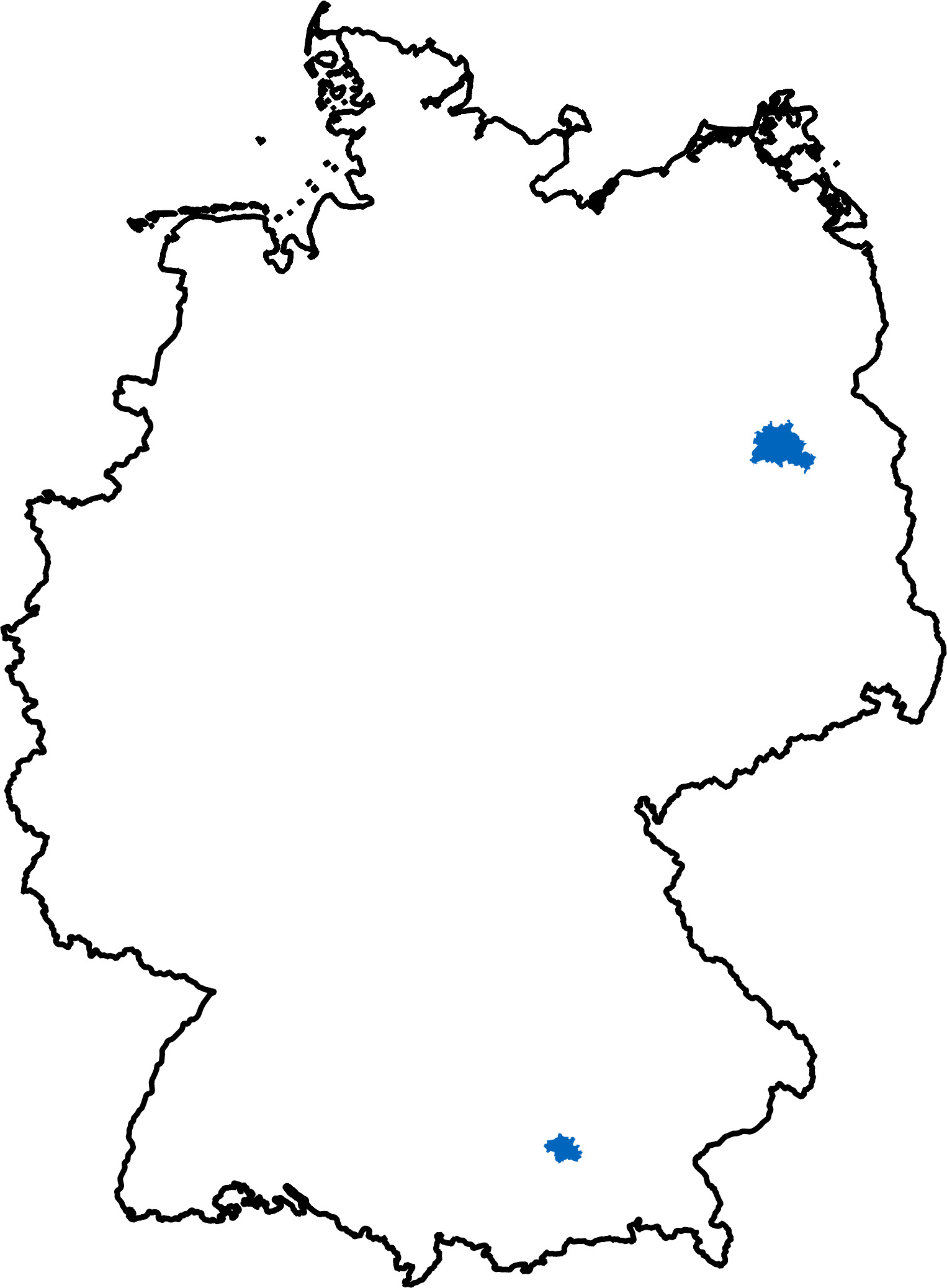}
        \caption{Location in Germany}
    \end{subfigure}
    \hfill
    \caption[]{Location of the considered study areas \tikz{\fill[orange, opacity=0.5] (0,0) rectangle (1.5ex,1.5ex);} within the densely built-up \tikz{\fill[black, opacity=.3] (0,0) rectangle (1.5ex,1.5ex);} cities of Berlin (a) and Munich (b) \tikz{\draw[thick] (0,0) rectangle (1.5ex,1.5ex);} that are located in the northern (Berlin) and southern (Munich) part of Germany (c).}
    \label{fig:study_area}
\end{figure}

Our experiments and evaluation were performed on two datasets acquired over study areas in the cities of Berlin and Munich, Germany.
The extent of both study areas is depicted in \cref{fig:study_area}.

\subsection{Data Processing}

\begin{figure}
  \begin{subfigure}[b]{.45\linewidth}
    \begin{tikzpicture}%
        \node [inner sep=0, outer sep=0] (a) at (0, 0) {\includegraphics[width=\linewidth]{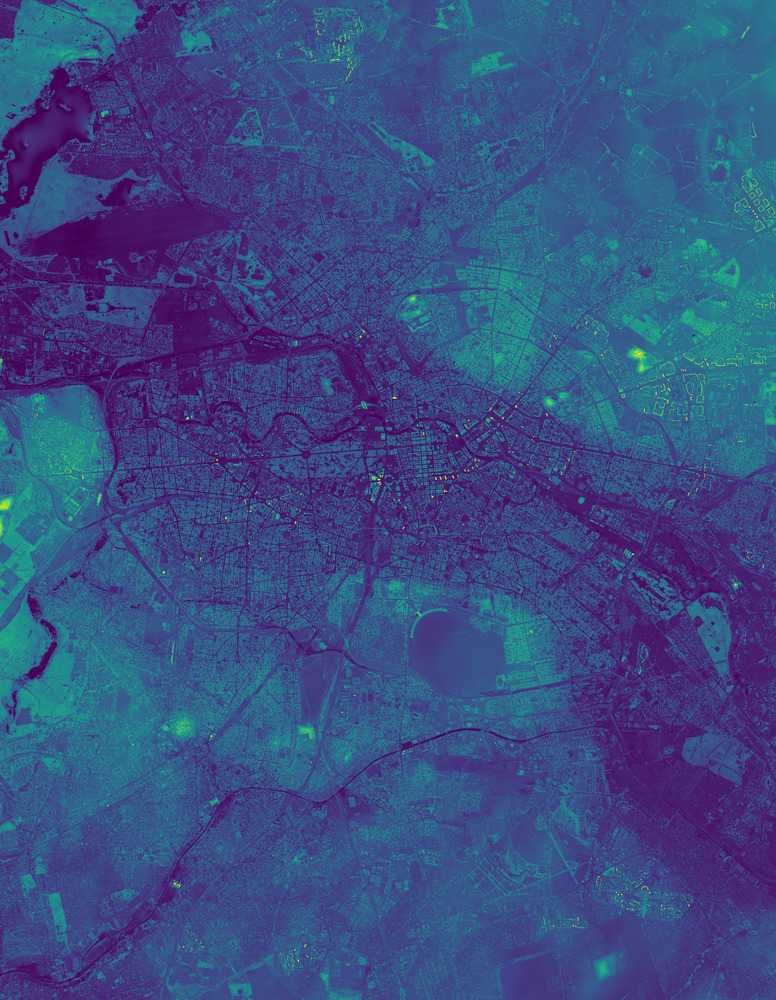}};
        \node [draw=white, line width=.5mm, above=0 of a.south east, anchor=south east, inner sep=0, outer sep=0] {\includegraphics[width=.5\linewidth]{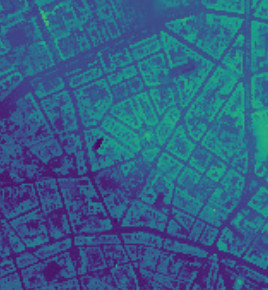}};
    \end{tikzpicture}
    \caption{Stereo DSM}
  \end{subfigure}
  \hfill
  \begin{subfigure}[b]{.45\linewidth}
    \begin{tikzpicture}%
        \node [inner sep=0, outer sep=0] (a) at (0, 0) {\includegraphics[width=\linewidth]{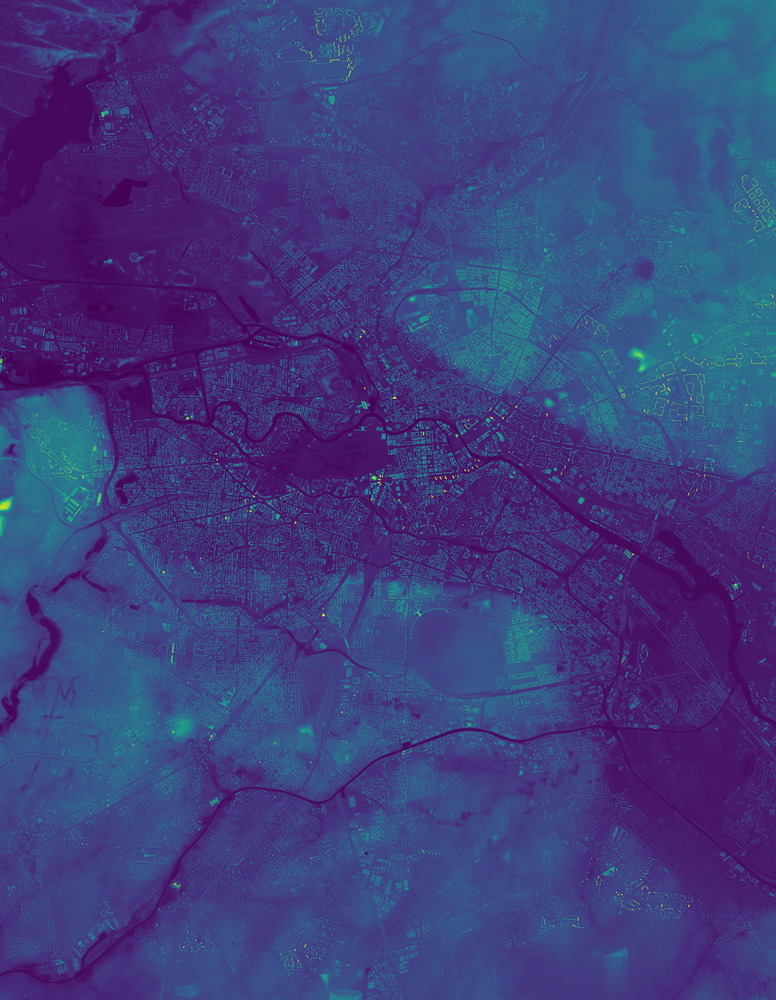}};
        \node [draw=white, line width=.5mm, above=0 of a.south east, anchor=south east, inner sep=0, outer sep=0] {\includegraphics[width=.5\linewidth]{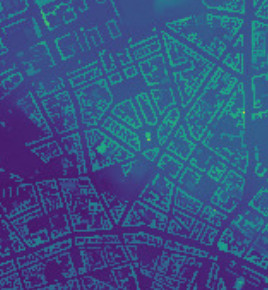}};
    \end{tikzpicture}
    \caption{Target DSM}
  \end{subfigure}
  \begin{subfigure}[b]{.5cm}
    \raggedleft
    \scriptsize
    high \\[.15cm]
    \rotatebox{90}{\scalebox{-1}[1]{\includegraphics[width=2.475cm, height=1ex]{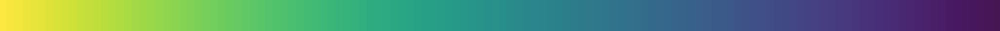}}} \rotatebox{90}{$\Delta_h \text{(clipped)} = \SI{75}{\meter}$} \\[.1cm]
    low
    \caption*{~}
  \end{subfigure}
  \caption{Overview and details of the raster elevation maps of Berlin consisting of a photogrammetric DSM (a) and target DSM derived from a DEM and a semantic 3D city model in LOD2 (b).}
  \label{fig:source_berlin}
\end{figure}

The Berlin dataset covers a total area of \SI{420}{\kilo\meter\squared}.
A photogrammetric \gls{dsm} with a rasterized \gls{gsd} of \SI{0.5}{\meter}, depicted in \cref{fig:source_berlin}, was generated using \gls{sgm} \citet{Hirschmueller08} on six pan-chromatic WorldView-1 images acquired on two different days, following the workflow of \citet{d2011semiglobal}.
The ground-truth data for the \gls{dsm} filtering task, \ie building geometries of a virtual city model in \gls{lod2} filled with a \gls{dem}, was obtained from a CityGML model freely available through the Berlin Open Data portal\footnote{\url{http://www.businesslocationcenter.de/downloadportal}}.
In order to convert the data to a suitable form, mainly the strategy introduced by \citet{bittner2018dsm} was applied.
The roof polygons from the CityGML data model were first triangulated using the algorithm proposed by \citet{shewchuk1996triangle} which is, in turn, based on Delaunay triangulation \citep{delaunay1934sphere}.
For the creation of a raster height map, a unique elevation value for each pixel inside the resulting triangles was calculated using barycentric interpolation.
Pixels that do not belong to a building were filled with \gls{dem} information.
The ground-truth data for the roof type segmentation task was computed from the previously obtained city model as well.
The slope for each pixel within the ground-truth height image was calculated as the maximum rate of elevation change between that pixel and its neighborhood.
The orientation of the computed slope was estimated clockwise from \SI{0}{\degree} to \SI{360}{\degree}, where north is represented by \SI{0}{\degree}, east by \SI{90}{\degree}, south by \SI{180}{\degree}, and west by \SI{270}{\degree}.
Finally, a ground-truth roof map with multiple classes was defined as follows: Class 0 corresponds to non-built-up areas, class 1 to flat roofs, and class 2 to sloped roofs.

\begin{figure}
  \begin{subfigure}{.45\linewidth}
    \includegraphics[width=\linewidth]{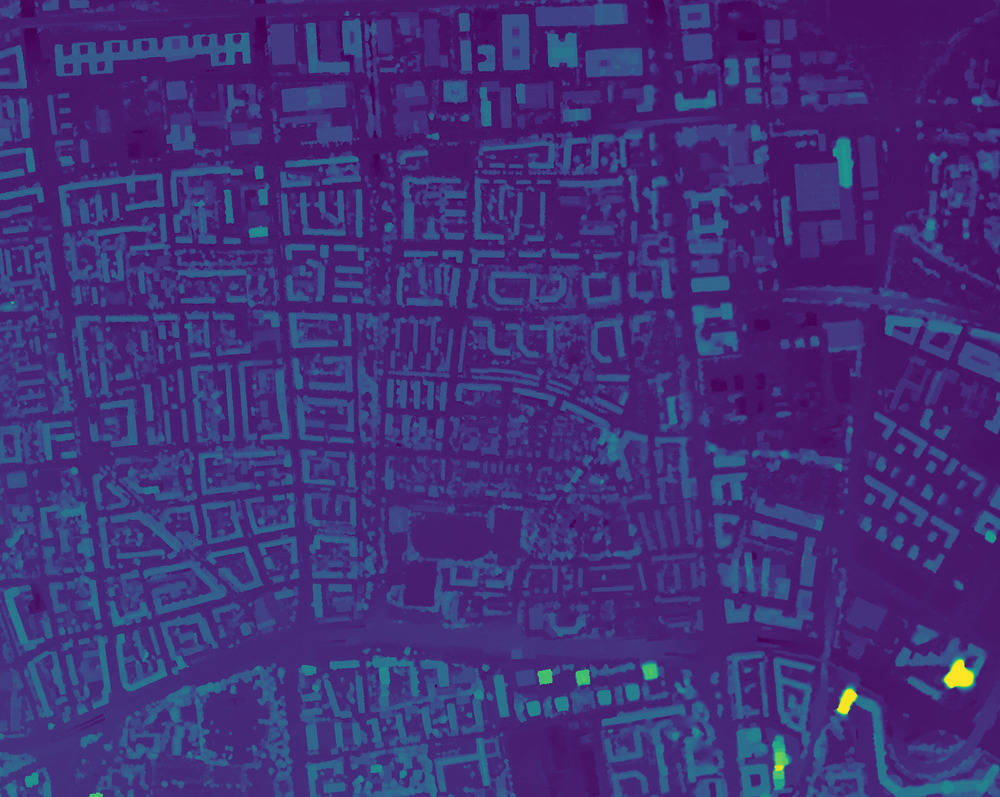}
    \caption{Stereo DSM}
  \end{subfigure}
  \hfill
  \begin{subfigure}{.45\linewidth}
    \includegraphics[width=\linewidth]{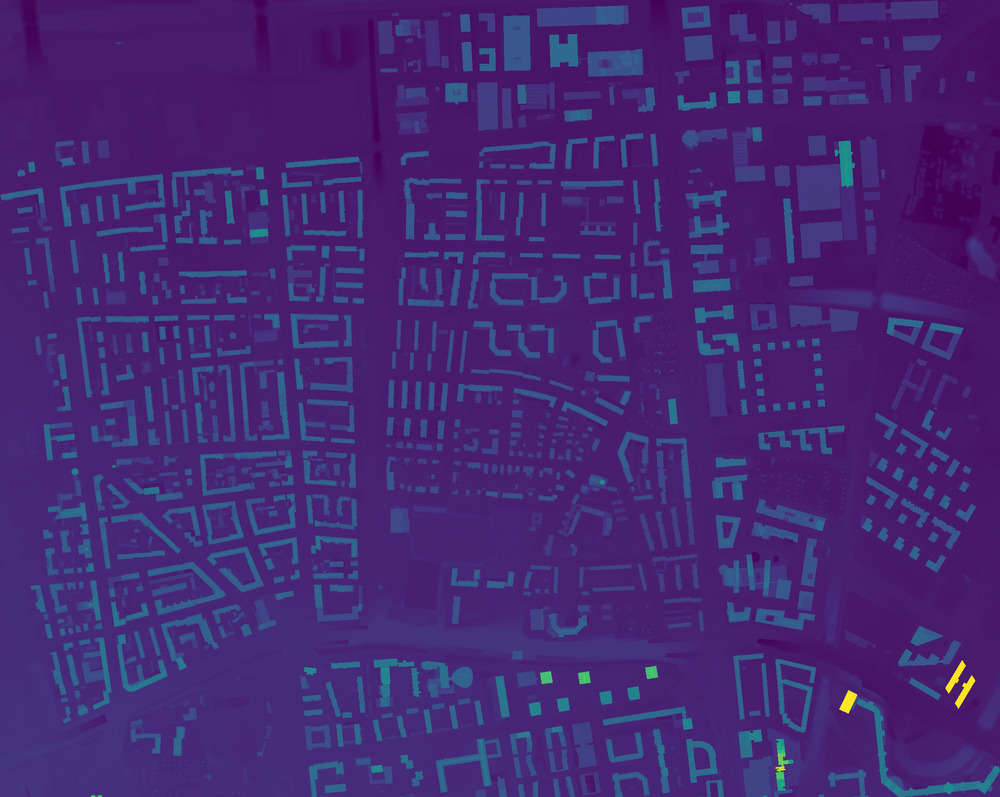}  
    \caption{Target DSM}
  \end{subfigure}
  \begin{subfigure}[c]{.5cm}
    \raggedleft
    \scriptsize
    high \\[.15cm]%
    \rotatebox{90}{\scalebox{-1}[1]{\includegraphics[width=2.475cm, height=1ex]{figures/data_source/colorbar_viridis}}} \rotatebox{90}{$\Delta_h \text{(clipped)} = \SI{75}{\meter}$} \\[.1cm]%
    low
    \caption*{~}
  \end{subfigure}

  \caption{Overview of the raster elevation maps of Munich consisting of a photogrammetric DSM (a) and target DSM derived from a DEM and a semantic 3D city model in LOD2 (b).
  This complementary test dataset is completely independent of the Berlin dataset, yet features similar properties.
  The temporal difference between the acquisition of source data for both products is clearly noticeable by new build and demolished buildings in the northwestern and eastern parts of the study area.}
  \label{fig:source_munich}
\end{figure}

The Munich study area covers a total area of \SI{3.8}{\kilo\meter\squared}.
A different space-borne sensor acquired the images used for generating the photogrammetric \gls{dsm} than in the Berlin study area.
Precisely, \gls{sgm} was applied on six pan-chromatic WorldView-2 images, which feature the same \gls{gsd} of \SI{0.5}{\meter} as the WorldView-1 images of Berlin to produce a \gls{dsm}, resampled to \SI{0.5}{\meter} \gls{gsd} accordingly.
Similarly to the other study area, the ground-truth for the target \gls{dsm} and the roof type mask were simulated from a CityGML model in \gls{lod2} and a \gls{dem}.
The virtual city model was provided by the Bavarian Agency for Digitization, High-Speed Internet and Surveying.
An illustration of the resulting height maps, given in \cref{fig:source_munich}, clearly shows the effect of a temporal offset in between the acquisition dates of the respective source data.
Especially noticeable are newly built and demolished buildings in the north western and eastern parts.

\subsection{Produced Datasets}

We split the Berlin dataset into regions for training, validation, and testing, as in prior work \citep{bittner2018dsm,bittner2019multi}.
These subsets are still closely related, since they originate from the same source.
Therefore, independent test data is required for evaluating the performance as to be expected in a real use-case.
We introduce an additional study area in Munich, which we exclusively use for testing selected models with a finalized set of network parameters, optimized on the training set, obtained with hyperparameters tuned on the validation set.

\begin{figure}
  \setlength{\fboxsep}{5pt}
  \setlength{\fboxrule}{.5pt}

  \centering

    \parbox{\linewidth-2\fboxsep}{
      \scriptsize
      \centering
      high \includegraphics[width=2cm, height=1.5ex]{figures/data_source/colorbar_viridis} low
      \hfill
      \tikz{\fill[tumblue] (0,0) rectangle (1.5ex, 1.5ex);} Flat Roof \quad
      \tikz{\fill[tumgreen] (0,0) rectangle (1.5ex, 1.5ex);} Non-Flat Roof \quad
      \tikz{\draw (0,0) rectangle (1.5ex, 1.5ex);} No Roof
    }%

  \vspace{2mm}

  \centering
  \setlength{\fboxsep}{0pt}

  \begin{subfigure}[c]{0.3\linewidth}
    \fbox{\includegraphics[width=\linewidth]{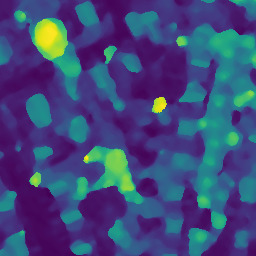}}
    \caption{Stereo DSM}
  \end{subfigure}
  \hfill
  \begin{subfigure}[c]{0.3\linewidth}
    \fbox{\includegraphics[width=\linewidth]{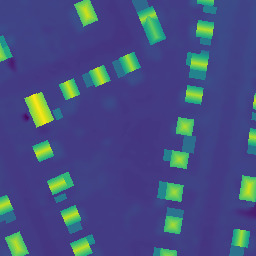}}
    \caption{Target DSM}
  \end{subfigure}
  \hfill
  \begin{subfigure}[c]{0.3\linewidth}
    \fbox{\includegraphics[width=\linewidth]{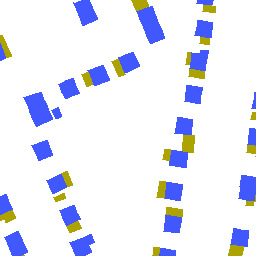}}
    \caption{Roof Types}
  \end{subfigure}

  \caption{Sample from our training dataset, with a patch, cropped from the photogrammetric DSM of Berlin (a) serving as input to the network, as well as ground-truth for the main task of DSM filtering (b) and roof type classification (c).}
  \label{fig:data_sample}
\end{figure}

During training, we cropped \SI{256x256}{\pixel} patches from the training area.
Each epoch consisted of a full randomized sweep over the training area.
As a data augmentation measure, we randomly shifted the original grid position of each patch by up to \SI{256}{\pixel}.
Hence, for each epoch, approximately \num{20000} samples were generated.
A sample from this dataset is illustrated in \cref{fig:data_sample}.

\section{Implementation and Experiments}
\label{sec:experiments}

In order to prove the feasibility and effectiveness of our proposed approach, we conducted numerous experiments and ablation studies to investigate the effect of each part in our system on the final result.
From the family of network architectures that can be generated using our generalized and modular multi-task concept, we selected and configured numerous instances that utilize different combinations of decoders and objective functions to gain insight into their suitability, both individually and in conjunction with each other.
We utilized renowned network architectures as building blocks within our modular framework.
Architectural considerations and respective evaluations of these networks and their specific features have been reported by the original authors and were re-validated by independent overview papers already.
Therefore, we refer the reader to the given references, to be found in the respective subsections, for in-depth analyses of the utilized network architectures.

\begin{figure*}
    \centering
    \includegraphics{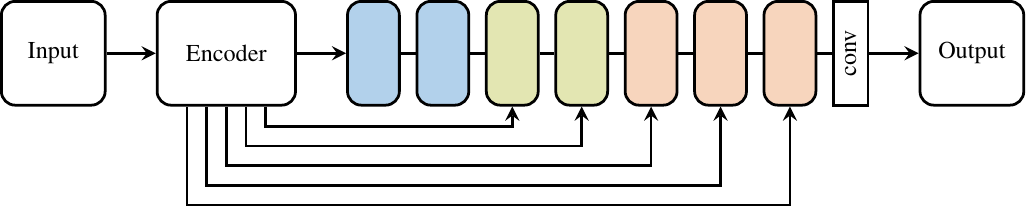}

    \vspace{.4cm}
    \includegraphics{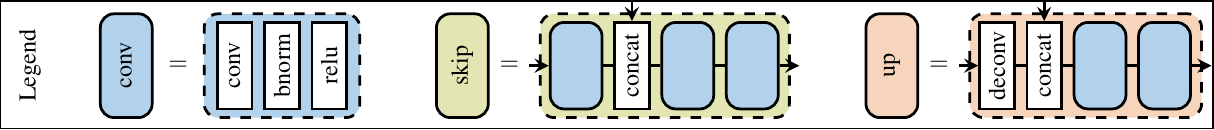}

    \caption{Architecture of a decoder module based on the decoder part of UNet.
    We replaced up-sampling blocks by simple skip blocks, to account for the resolution-preserving encoder employed in our experiments.
    With \num{137e6} learnable parameters this module outweighs the other considered decoders by far, resulting in the highest memory demand and inference time.}
    \label{fig:decoders_unet}
\end{figure*}

\begin{figure*}
    \centering
    \includegraphics{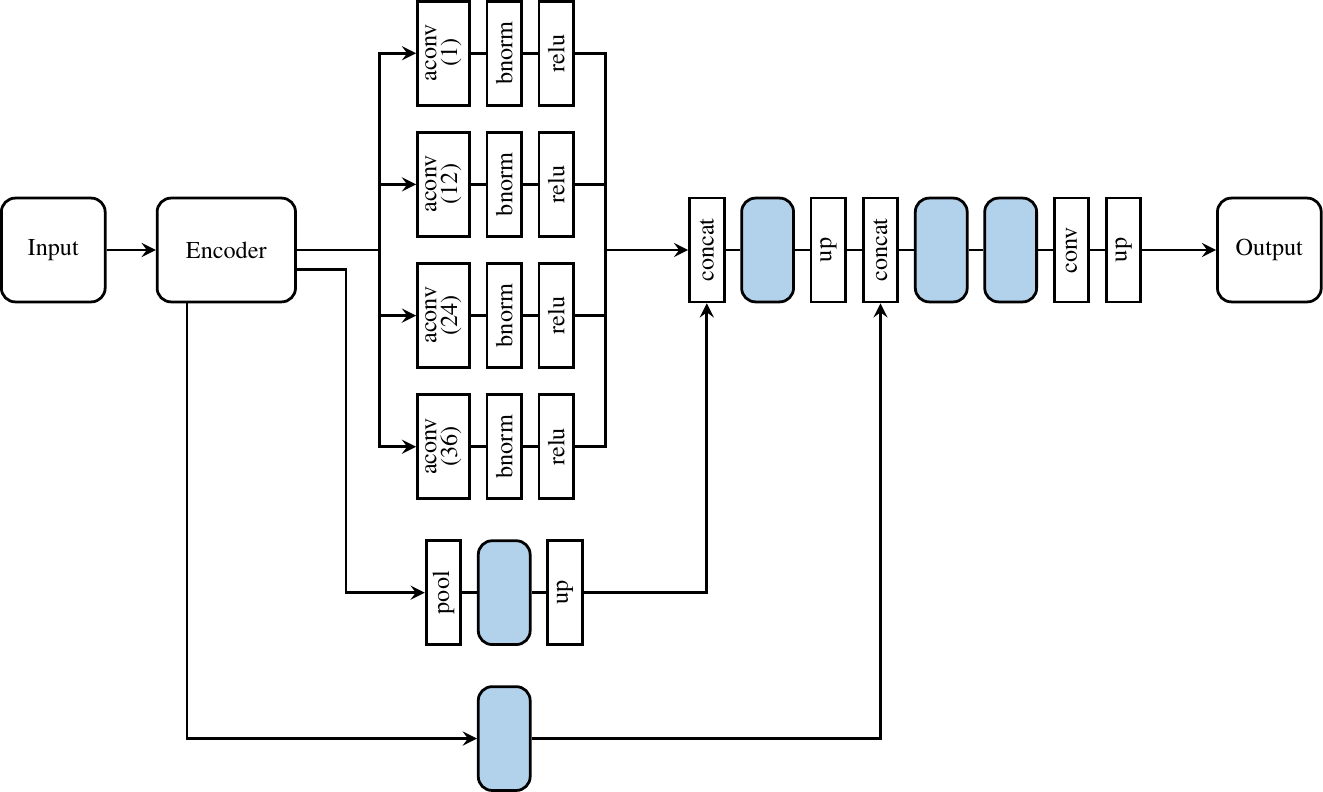}
    \caption{Architecture of a DeepLabv3+-based decoder as used in our experiments.
    This lightweight module with only \num{17e6} learnable parameters makes utilizes atrous spatial pyramid pooling.
    An output stride of eight is realized by the selected dilation rates in the atrous convolution layers.}
    \label{fig:decoders_deeplab}
\end{figure*}

\begin{figure*}
    \centering
    \includegraphics{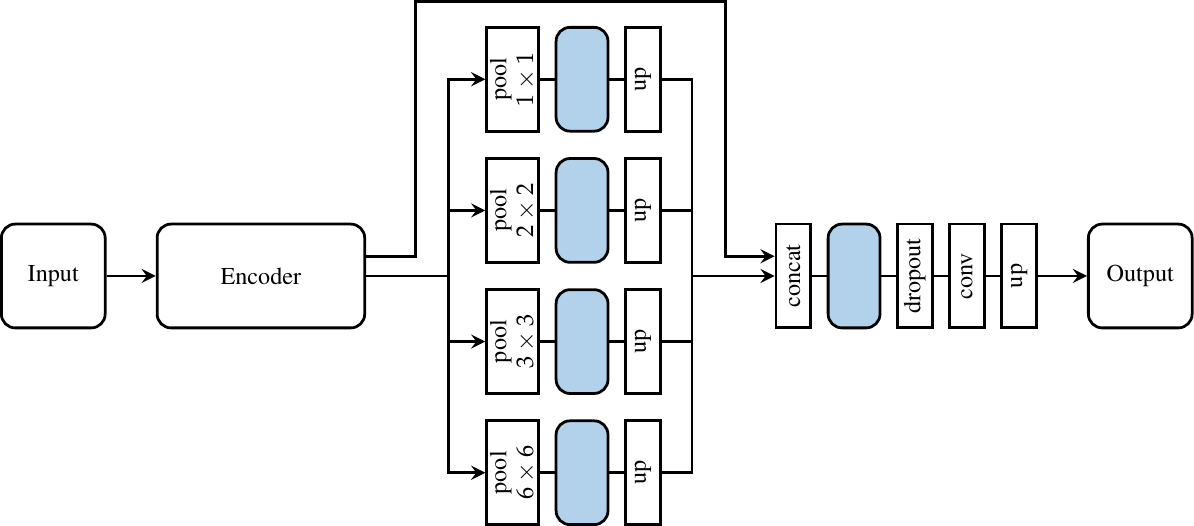}
    \caption{Architecture for a decoder module based on the PSPNet.
    The most prominent part of this module is its spatial pyramid pooling module that applies pooling with various kernel sizes in parallel.
    With \num{23e6} learnable parameters, the PSPNet decoder is similar in capacity to the DeepLabv3+ module and features the lowest computation time out of all considered decoders.}
    \label{fig:decoders_pspnet}
\end{figure*}

\subsection{The Encoder Module}
\label{subsec:encoders}

\Gls{cnn} architectures, developed for classification tasks, are widely studied and often employed as backbone networks in encoder-decoder systems for semantic segmentation.
Well-known examples for such architectures are the top-performing \gls{cnn} from the \gls{ilsvrc} \citep{ILSVRC15}, \ia Inception \citep{Szegedy15} and its successors, and various ResNet \citep{He16} variants that build upon the ideas of residual networks.
We mainly focus on ResNets, as they are versatile and can be implemented in various depths and providing skip features at different levels.
In order to retain the resolution of the input image, which is important for segmentation tasks, we modified the original architecture with atrous convolutions \citep{Chen15, Yu16}.
By employing a deep ResNet variant, namely an ImageNet pre-trained ResNet-101 with more than \num{42e6} parameters, as a backbone network, we ensure a sufficient capacity of the network.

\subsection{Decoder Modules}
\label{subsec:decoders}

Since both of our decoders are expected to produce a full-sized output map, the employed decoder architectures for the main and auxiliary tasks are interchangeable and equal---except for the very last layer that produces either one or three channels for the \gls{dsm} or roof type output, respectively.
All of the decoder modules we used in our experiments were derived from well-studied semantic segmentation networks and have successfully been re-utilized for regression tasks before.

\subsubsection{UNet}

A simple, yet powerful encoder-decoder based architecture called UNet was presented by \citet{Ronneberger15} for medical image analysis.
The network consists of symmetrical encoder and decoder sub-networks that are connected via skip connections.
Ever since the proposal of the original UNet, several variations have been proposed and applied in various works.
The original architecture consists of four down-sampling and respective up-sampling blocks and a bottleneck block in between.
For our decoder-only version though, we merely use the bottleneck and up-sampling parts of the UNet.
Since resolution-preserving encoders, such as the one employed in our experiments, retain a constant resolution of the feature maps up to a certain factor, we add two resolution-preserving skip blocks after the bottleneck block instead of up-sampling blocks.
They are, in fact, alike except for the very first layer, in which a standard convolution is applied instead of a deconvolution, as to be seen in \cref{fig:decoders_unet}.
The skip blocks are followed by three up-sampling blocks.
In total, the decoder takes advantage of five skip connections from low-level encoder features.
While this decoder uses rather simple building blocks, it comprises approximately \num{137e6} learnable parameters, which is by far the highest number of all decoders we implemented.

\subsubsection{DeepLabv3+}

The DeepLabv3+ \citep{Chen18} network achieved state-of-the-art results in multiple semantic segmentation benchmarks and challenges.
The architecture itself is a direct successor to DeepLabv3 \citep{Chen17}, which, in turn, descends from DeepLabv2 \citep{Chen18b} and the original DeepLab \citep{Chen15}.
Core concepts implemented in this family of \gls{cnn} architectures are atrous convolutions \citep{Chen15} and atrous spatial pyramid pooling \citep{Chen18b}.
Our DeepLabv3+ decoder, illustrated in \cref{fig:decoders_deeplab}, makes use of a single skip connection, similar to the original implementation.
Out of the three decoders employed in our experiments, the DeepLabv3+-based decoder with approximately \num{17e6} learnable parameters is the most lightweight.

\subsubsection{PSPNet}

Similar to DeepLabv3+, the PSPNet \citep{Zhao17} won several challenges for semantic segmentation by introducing effective techniques for pyramid scene parsing.
The signature component of this architecture is its pyramid pooling module that uses parallel pooling layers with different kernel sizes and subsequent concatenation of the resulting feature maps to capture scene context.
With approximately \num{23e6} learnable parameters, the capacity of our PSPNet-based decoder, illustrated in \cref{fig:decoders_pspnet}, is similar to the DeepLabv3+ variant.
As opposed to the other presented decoders, this one does not make use of any skip connections from low-level features of the encoder.

\begin{figure*}[t!]
    \centering
    \includegraphics{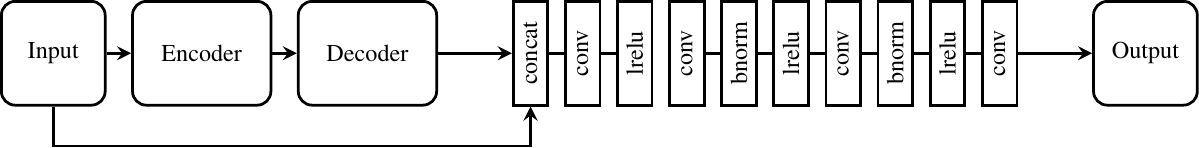}
    \caption{Architecture of the conditional discriminator network used in our experiments.
    This fully-convolutional PatchGAN-based discriminator is conditioned on the initial stereo DSM and assigns binary labels to fixed-sized image patches for arbitrarily large inputs.
    A simple MSE loss serves as an objective function for training.
    }
    \label{fig:discriminator}
\end{figure*}

\subsection{Objectives}
\label{subsec:objectives}

In order to train our model for the desired main goal of \gls{dsm} filtering, we defined multiple objectives.
In addition to an \lone{} reconstruction loss, we evaluated surface normals to encourage the prediction of smooth surfaces.
Since reconstruction losses favor slightly blurry predictions, we added a conditional adversarial loss to force the network to predict crisp boundaries.
All of the defined objective functions have to be combined into a single loss function to facilitate optimization.
By balancing their contribution with corresponding weighting terms, we account for the different properties of each of those.
Implementation details are described in the following.

\subsubsection{Reconstruction Loss}

A basic reconstruction loss was applied to evaluate the quality of the estimated \gls{dsm}.
We implemented an \lone{} loss \loneloss{} that favours clear and detailed predictions rather than oversmoothed results, as often observed when applying \ltwo{} losses \citep{Carvalho18}.
While this objective is rather basic, it is expected to contribute to the final result the most due to its unique property of assessing absolute height errors.
Therefore, we treated this objective as the core part of our multi-objective multi-task loss function and utilized it in all of our experiments.
To further boost the quality of predictions, we added more specialized objectives to account for specific shortcomings of this reconstruction loss.

\subsubsection{Surface Normal Loss}

Since smoothness and crisp boundaries are not explicitly enforced by the basic reconstruction loss, we add the evaluation of surface normals to our set of objectives.
We derive normal directions from gradients in the image representation of the height maps \citep{Hu19}.
Hence, this objective makes use of the same predictions as the reconstruction loss and, thus, does not require an individual decoder.
The employed surface normal loss
\begin{align}
  \loss_\text{n} = 1 - \left ( \frac{1}{N} \sum_{i = 1}^N \left ( \frac{\V{n}_i^\top\V{n}_i^*}{\lVert \V{n}_i \rVert \cdot \lVert \V{n}_i^*\rVert} \right ) \right )
\end{align}
evaluates the remaining angular difference between the normal directions in prediction $\V{n}_i$ and ground-truth $\V{n}_i^*$, normalized to unit length.

\subsubsection{Conditional Adversarial Loss}


While typical reconstruction losses favor slightly blurry predictions to avoid larger errors, approaches making use of adversarial losses have been shown to produce sharper edges in recent work \citep{Carvalho18, bittner2019multi} since such systematical errors are easy for a discriminator to detect.
We implemented a \gls{cgan}-based objective through a conditional discriminator network $\mathfrak{D}$.
Since deep high-capacity discriminator network architectures add numerous parameters to the optimization problem but only improved the quality of results negligibly in preliminary experiments, we employed a simple PatchGAN discriminator architecture \citep{Isola17} that classifies \SI{70x70}{\pixel} patches instead of full images.
It is, therefore, able to classify arbitrarily sized images while only using approximately \num{2.8e6} parameters. 
We modified the original architecture slightly by feeding in concatenated predictions and input images, as shown in \cref{fig:discriminator}, thus making it a conditional discriminator.
$\mathfrak{D}$ operates on the input stereo \gls{dsm} $\V{x}$ and a filtered \gls{dsm} $\V{y}_\text{DSM}$ or $\V{y}^*_\text{DSM}$ and assigns patch-wise labels $\{0: \text{fake}, 1: \text{real}\}$.
During training, we alternately optimized the main encoder-decoder network and the discriminator network using a simple \gls{mse} loss.
We derive our objective function $\loss_\text{GAN}$ from the discriminator decision as
\begin{align}
  \loss_\text{GAN} = \frac{1}{N} \sum_{i = 1}^N \left ( \mathfrak{D}(\V{x}_i, \V{y}_{\text{DSM}, i}) - 1 \right )^2 \quad .
\end{align}
Thus, $\loss_\text{GAN}$ yields low values when the discriminator network mistakes predictions for ground-truth samples.

\subsubsection{Segmentation Loss}

The main and only objective we employed for roof type segmentation was a standard SoftMax cross-entropy loss
\begin{align}
    \loss_\text{seg} = \frac{1}{N} \sum_i^N - y_{\text{seg}, i, y^*_{\text{seg}, i}} + \log \left( \sum_{c \in \Set{c}} y_{\text{seg}, c} \right)
\end{align}
that evaluates per pixel predictions $\M{y}_\text{seg} = \VecDef{y_{\text{seg}, i, c}}_{i, c}$ with target labels $\V{y}^*_\text{seg} = \VecDef{y^*_{\text{seg}, i}}_{i}$ for the three-class segmentation task.
This objective was expected to mainly provide supervision for building footprint recognition in order to preserve and refine the shape of buildings in the predicted \gls{dsm}.
As a secondary source of information, buildings are classified based on their roof geometries with classes for flat and non-flat roofs.
In the closely related field of depth estimation it can be observed that the optimization of regression tasks is generally harder than the optimization of corresponding classification tasks \citep{Fu18, Liebel19}.
By restricting the highly non-convex optimization space through this constraint, this auxiliary objective doubles as a regularization measure \citep{Liebel18}.

\subsection{Multi-Objective and Multi-Task Loss Function}

The set of objective functions to be utilized during training a specific model has to be incorporated into a single multi-objective and multi-task loss function, subject to optimization.
We applied weighting based on homoscedastic task uncertainty $\sigma_\tau$\citep{Kendall18}, treating each objective as a \enquote{task} $\tau$.
As per \cref{eq:kendall_loss}, we designed our multi-task loss for the regression objectives $\loneloss$ and $\loss_{n}$, the classification objective $\loss_\text{seg}$, and the conditional adversarial objective $\loss_\text{GAN}$---for which we fix the weight---as
\begin{align}
  \loss_\text{mt} &= \sum_\tau \loss_\tau \cdot w_\tau + r_\tau \\[2mm]
                  &= \left(\loneloss        \cdot w_{\lone}      + r_{\lone}       \right)
                   + \left(\loss_\text{n}   \cdot w_{\text{n}}   + r_{\text{n}}    \right) \nonumber \\
                   &\quad+ \left(\loss_\text{seg} \cdot w_{\text{seg}} + r_{\text{seg}}  \right)
                   + \loss_\text{GAN} \cdot w_{\text{GAN}} \\[5mm]
                  &= \left( \loneloss              \cdot \frac{\exp(-s_{\lone})}{2}    + \frac{s_{\lone}}{2} \right) \nonumber \\
                   &\quad+ \left( \loss_\text{n}         \cdot \frac{\exp(-s_{\text{n}})}{2} + \frac{s_{\text{n}}}{2} \right) \nonumber \\
                   &\quad+ \left( \loss_\text{seg} \cdot \exp(-s_{\text{seg}})         + \frac{s_{\text{seg}}}{2} \right) \nonumber \\
                   &\quad+ \loss_\text{GAN}       \cdot s_{\text{GAN}} \label{eq:mtloss} \\[5mm]
                  &= \left( \loneloss              \cdot \frac{\exp(-\log ( \sigma_{\lone}^2 ))}{2}    + \frac{\log ( \sigma_{\lone}^2 )}{2} \right) \nonumber \\
                   &\quad+ \left( \loss_\text{n}         \cdot \frac{\exp(-\log ( \sigma_\text{n}^2 ))}{2}   + \frac{\log ( \sigma_\text{n}^2 )}{2} \right) \nonumber \\
                   &\quad+ \left( \loss_\text{seg} \cdot \exp(-\log ( \sigma_\text{seg}^2 ))           + \frac{\log ( \sigma_\text{seg}^2 )}{2} \right) \nonumber \\
                   &\quad+ \loss_\text{GAN}       \cdot \log ( \sigma_\text{GAN}^2 ) \\[5mm]
                  &= \frac{1}{2} \left( \frac{\loneloss}{\sigma_{\lone}^2} + \log ( \sigma_{\lone}^2 ) \right)
                   + \frac{1}{2} \left( \frac{\loss_\text{n}}{\sigma_\text{n}^2} + \log ( \sigma_\text{n}^2 ) \right)  \nonumber \\
                   &\quad+ \left( \frac{\loss_\text{seg}}{\sigma_\text{seg}^2} + \frac{\log ( \sigma_\text{seg}^2 )}{2} \right)  \nonumber \\
                   &\quad+ \loss_\text{GAN}       \cdot \log ( \sigma_\text{GAN}^2 )
                  ~ \text{.}
\end{align}
The learned weighting terms represent a relative balancing of the tasks, thus one of them can safely be fixed.
Fixing the weight for $\loss_\text{GAN}$ loss lead to more stable convergence in preliminary experiments, since it is a combination of multiple losses itself and thus neither satisfies the underlying assumptions for classification nor regression losses.
As a consequence, no regularization value $r_\text{GAN}$ is required.
Note that we optimized for $s_\tau := \log \sigma_\tau^2$, as given in \cref{eq:mtloss}, instead of $\sigma_\tau$ due to higher numerical stability.

\begin{figure}
  \centering
  \includegraphics{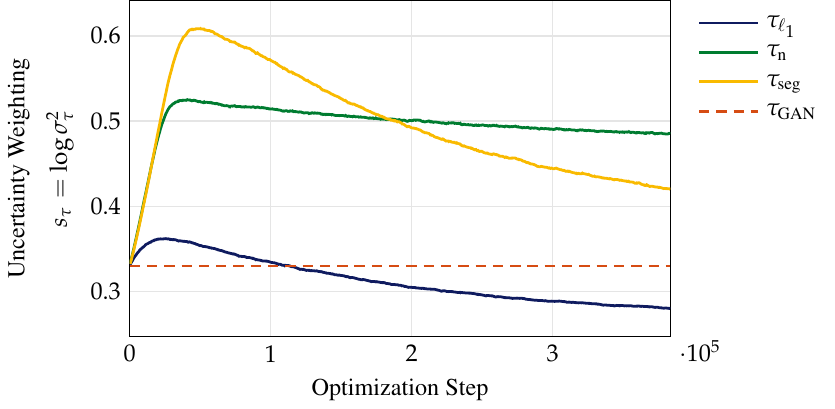}
  \caption{Evolution of task balancing weights during the training phase of one of our models with a multi-task DeepLabv3+/PSPNet configuration.
  The applied weighting scheme is based on an uncertainty measure, represented by $\sigma_\tau$, which were optimized along with the network parameters.
  Since this measure does not directly apply to adversarial terms, we fixed the corresponding weight to its initial value.}
  \label{fig:task_weights}
\end{figure}

\subsection{Experimental Design}

Based on the various architectural modules and objective functions presented in \cref{subsec:encoders,subsec:decoders,subsec:objectives}, we conducted experiments using different instances of our generalized framework with various combinations of network architectures and objectives.

In order to restrict the study to a feasible number of experiments, we alternately fixed parts of the system while validating a broad range of configurations for the other parts.
We started by evaluating the impact of each objective on the overall performance.
For the best performing candidate, we ran further experiments with a variety of decoder architectures.

By training and validating each network on the respective subset of the Berlin study area, we make the results comparable while leaving out the Berlin and Munich test areas to avoid overfitting.
Only the best performing models from the validation stage were subsequently evaluated on the test sets.
This independent testing stage, thus, gives an impression of the generalization performance, \ie performance on unseen data as in a real-world use-case.
A more detailed description of the testing procedure along with the respective results is given in \cref{sec:evaluation}.

\subsection{Training Settings}

For running the experiments, our proposed approach was implemented in PyTorch.
Keeping the hyperparameter settings consistent for all experiments allowed for comparing the results directly.
We used batches of size $N_\text{batch} = 5$, a fixed learning rate $\alpha = 0.0005$ for the Adam optimizer \citep{Kingma15} with $\beta_1 = 0.9, \beta_2 = 0.999$, and ran training for a fixed number of $N_\text{epoch} = 100$ epochs on the Berlin training subset, which roughly equals $N_\text{iter} = \num{38e4}$ optimization steps.
Training was conducted on a single NVIDIA 1080Ti GPU.
We selected the final set of parameters based on validation metrics calculated during training from the validation subset.
Weighting terms for the contributing objectives were dynamically adjusted with the network parameters.
\Cref{fig:task_weights} shows the development of these parameters over the course of a full training run.
While we fixed $s_\text{GAN}$ to its initial value, the remaining weights were quickly adjusted to the right scale automatically.
The task weights slowly decayed from this temporary high, mirroring the overall increase in performance of the network.

\subsection{Validation of Training Results}

The modular approach of our implementation with interchangeable encoder and decoder architectures, and multiple choices for objective functions, allows for numerous unique configurations.
Out of this pool of options, we picked various relevant configurations and trained them according to the aforementioned procedure and datasets.

Results were evaluated using standard metrics for both tasks.
We mainly measured the quality of the filtered \gls{dsm} using the \gls{rmse}
\begin{align}
    \text{RMSE}&(\V{y}_\text{DSM}, \V{y}^*_\text{DSM}) = \nonumber \\
    & \frac{1}{N} \sqrt{( \V{y}_\text{DSM} - \V{y}_\text{DSM}^* )^\top \cdot ( \V{y}_\text{DSM} - \V{y}_\text{DSM}^* )}
\end{align}
with predictions $\V{y}_\text{DSM}^*$ and ground-truth labels $\V{y}_\text{DSM}$.
In addition to that, we evaluated the \gls{mae}
\begin{align}
    \text{MAE}(\V{y}_\text{DSM}, \V{y}^*_\text{DSM}) = \frac{1}{N} \sum_{i = 1}^N | y_{\text{DSM}, i} - y^*_{\text{DSM}, i} | \quad \text{.}
\end{align}
All of these metrics are given in meters, where lower values correspond to lower errors and, thus, better performance.
Similar to these standard metrics for regression results, we use the common \gls{miou} metric
\begin{align}
    \text{mIOU}(\V{y}_\text{seg}, \V{y}_\text{seg}^*; &\Set{c}) = \nonumber \\
    &\frac{1}{|\Set{c}|} \sum_{c \in \Set{c}} \frac{| \Set{R}_c(\V{y}_\text{seg}) \cap \Set{R}_c(\V{y}^*_\text{seg}) |}{| \Set{R}_c(\V{y}_\text{seg}) \cup \Set{R}_c(\V{y}^*_\text{seg}) |}
\end{align}
for validating the segmentation performance of the auxiliary task with three classes $\Set{c}$ that represent $\{\text{no building}, \text{flat roof}, \text{sloped roof}\}$, with higher values representing better performance.

\subsubsection{Multi-Objective Ablation Study}

\begin{table}
    \vspace{-.6\baselineskip}
    \caption{Results of an ablation study on the Berlin validation set showing the training time and performance of instances of our model trained with an increasing number of objectives.
    In the case of multi-objective training the utilized weighting scheme is given.
    Best results are highlighted in blue color.
    Multi-task training with learned weighting terms yields best results, both in terms of RMSE and required training time.}
    \label{tab:results_val_ablation}

    \begin{tabularx}{\linewidth}{@{}*{5}{l}S[table-format=2.0, table-number-alignment=right]S[table-format=1.2, table-number-alignment=right]@{}}

        \toprule

        \multicolumn{4}{@{}l}{\hf Objectives} & \multicolumn{1}{l}{\hf Weighting} & \multicolumn{2}{l@{}}{\hf Performance} \\

        \cmidrule(r){1-4} \cmidrule(lr){5-5} \cmidrule(l){6-7}

        \multicolumn{1}{@{}X}{\hf \loneloss} & \multicolumn{1}{X}{\hf $\mathcal{L}_\text{n}$} & \multicolumn{1}{X}{\hf $\mathcal{L}_\text{GAN}$} & \multicolumn{1}{X}{\hf $\mathcal{L}_\text{seg}$} & & \multicolumn{1}{l}{\hf Best Epoch $\uparrow$} & \multicolumn{1}{l@{}}{\hf RMSE (in m) $\uparrow$} \\

        \cmidrule(r){1-1} \cmidrule(lr){2-2} \cmidrule(lr){3-3} \cmidrule(lr){4-4} \cmidrule(lr){6-6} \cmidrule(l){7-7}

        $\boxtimes$ & $\Box$      & $\Box$      & $\Box$        & \hf \textemdash & 96 & 1.81 \\
        $\boxtimes$ & $\boxtimes$ & $\Box$      & $\Box$        & \hf learned & 72 & 1.69 \\
        $\boxtimes$ & $\boxtimes$ & $\boxtimes$ & $\Box$        & \hf learned & 78 & 1.68 \\
        $\boxtimes$ & $\boxtimes$ & $\boxtimes$ & $\boxtimes$   & \hf fixed   & 83 & \color{tumblue}1.67 \\
        $\boxtimes$ & $\boxtimes$ & $\boxtimes$ & $\boxtimes$   & \hf learned & \color{tumblue}59 & \color{tumblue}1.67 \\

        \bottomrule

    \end{tabularx}

\end{table}

One of the most important aspects of this evaluation is to re-validate the multi-task multi-objective approach in an ablation study.
Hence, we trained a family of networks that utilize a gradually increasing number and combination of objectives and tasks.
We considered configurations from the most basic single-task and single-objective \loneloss{} \gls{dsm} filtering network up to a full-fledged version using the complete set of implemented objectives and tasks.
We used a UNet decoder for the main task and a DeepLabv3+ decoder for the segmentation task in these experiments.
The validation results, given in \cref{tab:results_val_ablation}, show that each objective that was added to the loss function improved the predictive performance on unseen data.
Consequently, additional supervision through the secondary task and its respective objective function $\loss_\text{seg}$ was expected to boost the performance further.
Our experimental results show that multi-task training not only improved the performance in terms of \gls{rmse} by scoring the lowest error but also dramatically decreased training time.
While best results were achieved after epoch \num{96} using single-objective training, and after epochs \num{72} and \num{78} using multi-objective training, the multi-task multi-objective variant of the network peaked performance after epoch \num{59} already.
Hence, by making use of the proposed method with the full set of implemented objectives, roughly \SI{40}{\percent} of training time could be saved, adding training time efficiency to its advantages.
Another important factor in this is the employed weighting scheme, as to be seen from the results for multi-task multi-objective training with equal weights that required 83 epochs of training to achieve comparable performance.

\subsubsection{Model Selection}

\begin{figure}
  \centering
  \includegraphics{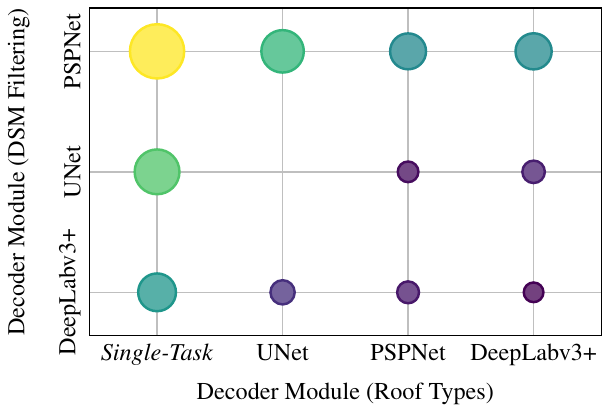}
  \includegraphics{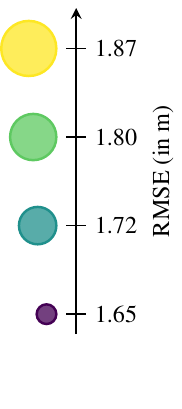}

  \caption{Performance of selected instances of our model utilizing various decoders for the main DSM filtering task and the secondary roof type classification task.
  Single-task and single-objective training with \loneloss{} exclusively is given as a baseline reference for each DSM filtering decoder architecture.}
  \label{fig:val_results_decoder_architectures}
\end{figure}

We deem it crucial to stick to a strict three-fold evaluation scheme with independent training, validation, and testing phases.
Hence, we selected the best models and procedure for the final testing stage from the validation results.
To this end, we compared the performance of multi-task models that follow the results of the prior ablation study (\cf \cref{tab:results_val_ablation}), \ie the full set of implemented objectives and tasks.
\Cref{fig:val_results_decoder_architectures} illustrates the results of the validation stage for models utilizing various combinations of decoder architectures, with \loneloss{} optimization serving as a single-task single-objective baseline.
It became apparent that there is a clear order in the suitability of decoder architectures for \gls{dsm} filtering according to the baseline results, with the DeepLabv3+ decoder scoring the lowest and the PSPNet decoder the highest \gls{rmse}.
Similarly, DeepLabv3+ and PSPNet decoders consistently achieve better performance in roof type classification than the UNet decoder.
The difference in \gls{rmse} between the configurations using DeepLabv3+ and UNet decoders for the main task combined with all three decoders for the auxiliary task is negligible.
Hence, all five of these configurations were advanced to the testing stage.
Note that due to technical limitations, experiments for the UNet/UNet configuration could not be conducted using a similar setup as the other ones due to extensive memory demand.
We intentionally refrained from making individual adjustments in the training settings for this specific configuration in order to retain a maximum of comparability.

\section{Evaluation}
\label{sec:evaluation}

For the final testing stage, we prepared a test set from each of our two study areas.
While the Berlin area was split into training, validation, and test as described in \cref{sec:data}, the Munich dataset was exclusively used for testing.
Only the best-performing models on the validation set were considered in the test phase in order to avoid overfitting on the test split.
In order to be able to set the results into perspective, we implemented a baseline that utilizes hand-crafted geometric features, described in \cref{subsec:baseline}.
Results for the Berlin study area are presented in \cref{subsec:results_berlin}.
The completely independent Munich test set gives further insight into the generalization performance of the proposed method.
Test results for this additional evaluation stage are given in \cref{subsec:results_munich}.

The test areas were split into overlapping patches of \SI{256x256}{\pixel}.
A stride of \SI{64}{\pixel} was implemented in order to reduce the influence of border effects by averaging over multiple predictions.
While this measure increases the total number of inference steps, it helps to generate smooth predictions over the whole test region.
Note that the proposed network architectures are fully convolutional and---given sufficient memory---therefore, able to generate predictions of arbitrary sizes in a single inference step in theory.

A quantitative evaluation of the results was conducted using the \gls{rmse} as our main metric, similar to the validation stage.
We also evaluate the \gls{miou} of the predicted roof type maps, even though they merely serve as a regularization measure through additional supervision and are not a primary product of our system.

\begin{table}
  \vspace{-.6\baselineskip}
  \caption{Evaluation results for multiple instances of our multi-task framework on the Berlin test set.
  As expected, all methods easily outperformed the very basic stereo DSM baseline (first line), which serves as an input to the networks, as well as the stronger baseline using hand-crafted geometric features (second line).
  All variants of our model (lines 4-8) also considerably outperformed the state of the art (third line) in all metrics.
  Best performance (highlighted in blue) was achieved by fusing the prediction results of the five evaluated instances of our model using a simple ensemble approach.
  }
  \label{tab:test_results_berlin}

  \begin{tabularx}{\linewidth}{@{}l*{2}{S[table-format=1.4, table-number-alignment=right]}S[table-format=2.2, table-number-alignment=right]@{}}
    \toprule
    \multicolumn{1}{@{}X}{\hf Method} & \multicolumn{1}{l}{\hf RMSE $\uparrow$} & \multicolumn{1}{l}{\hf MAE $\uparrow$} & \multicolumn{1}{l@{}}{ \hf mIOU $\downarrow$} \\[-1.5mm]
    & \multicolumn{1}{l}{\hf (in m)} & \multicolumn{1}{l}{\hf (in m)} & \\
    \cmidrule(r){1-1} \cmidrule(lr){2-2} \cmidrule(lr){3-3} \cmidrule(l){4-4}
    \hf Stereo DSM                        & 7.14 & 1.99 & \textemdash \\
    \hf Baseline                          & 5.99 & 1.69 & \textemdash \\
    \hf \Citet{bittner2018dsm}            & 2.95 & 1.20 & \SI{64.02}{\percent} \\[1mm]
    \hf Ours (UNet/PSPNet)                & 2.17 & 1.07 & \SI{64.50}{\percent} \\
    \hf Ours (DeepLabv3+/DeepLabv3+)      & 2.13 & 1.11 & \SI{64.73}{\percent} \\
    \hf Ours (UNet/DeepLabv3+)            & 2.04 & 1.06 & \SI{64.14}{\percent} \\
    \hf Ours (DeepLabv3+/UNet)            & 2.04 & 1.10 & \SI{64.48}{\percent} \\
    \hf Ours (DeepLabv3+/PSPNet)          & 2.01 & 1.09 & \SI{64.90}{\percent} \\[1mm]
    \hf \color{tumblue}{Ours (Ensemble)}  & \color{tumblue}1.94 & \color{tumblue}1.06 & \color{tumblue}\SI{65.85}{\percent} \\
    \bottomrule
  \end{tabularx}
\end{table}

\subsection{Baseline Method}
\label{subsec:baseline}

Multiple solutions have been proposed for distinguishing object from ground points and subsequent removal of such for the purpose of deriving digital terrain models from \glspl{dsm} \citep{Tovari05,Jacobsen03,Salah20}.
While these methods are closely related to the desired removal of vegetation, they also target buildings and other structures.
In contrast to focusing on removing a certain class of objects, they often detect points to keep and interpolate between them.

\citet{Anders19} conducted a comprehensive comparison of various methods for removing vegetation from \gls{uav}-based photogrammetric point clouds.
The presented methods are applicable to our task in general, however, they operate on point clouds instead of a raster DSM.
Hence, utilizing such approaches would require a transformation from height maps to point clouds.
While this is feasible, the characteristics of the resulting product would differ from photogrammetric point clouds.
Most notably, a transformed point cloud will exhibit low point density due to the prior lossy rasterization process.

Considering these shortcomings, we implemented a more classical method based on manually designed geometric features as a baseline.
Utilizing a straightforward and hand-crafted approach as a baseline comes with the major benefit of being easily interpretable.
Following \citet{Weidner97}, we expect vegetation to be detectable in the \gls{dsm} from a high variance in the direction of surface normals.
Hence, we extracted surface normals from the height map and calculated their variance in a defined neighborhood around each pixel.
A segmentation map was derived from this by applying a threshold.
To reduce noise in the resulting binary mask, we employed morphological opening and closing.
The masked pixels were removed from the \gls{dsm}.
Missing values were subsequently interpolated.
We considered various interpolation methods out of which filling with the minimal value in a specified window around the masked pixel yielded the best results.
The parameters and interpolation method for this algorithm were manually tuned on samples from the training set and applied to the test areas without further changes.

\subsection{Test Results}
\label{subsec:results_berlin}

Test results for the Berlin study area, given in \cref{tab:test_results_berlin}, show that the proposed method is able to filter stereo \glspl{dsm}.
The residual between the input \gls{dsm} and the ground-truth with an \gls{rmse} of \SI{7.14}{\meter} can be decreased to \SI{5.99}{\meter} by the baseline method.
The current state of the art, which achieves an \gls{rmse} of \SI{2.95}{\meter}, further improves on this.
However, all of the tested architectural variants of the proposed method were able to outperform the state of the art.
With an \gls{rmse} of \SI{2.01}{\meter}, one instance of the proposed method, implemented using decoders based on DeepLabv3+ and PSPNet for the main DSM filtering and roof type classification task, respectively, scored the lowest error.

\begin{figure}
  \setlength{\fboxsep}{0pt}%
  \setlength{\fboxrule}{.5pt}%
  \begin{subfigure}{.23\linewidth}
    \fbox{\rotatebox{90}{\includegraphics[height=\linewidth, trim=30 0 120 0, clip]{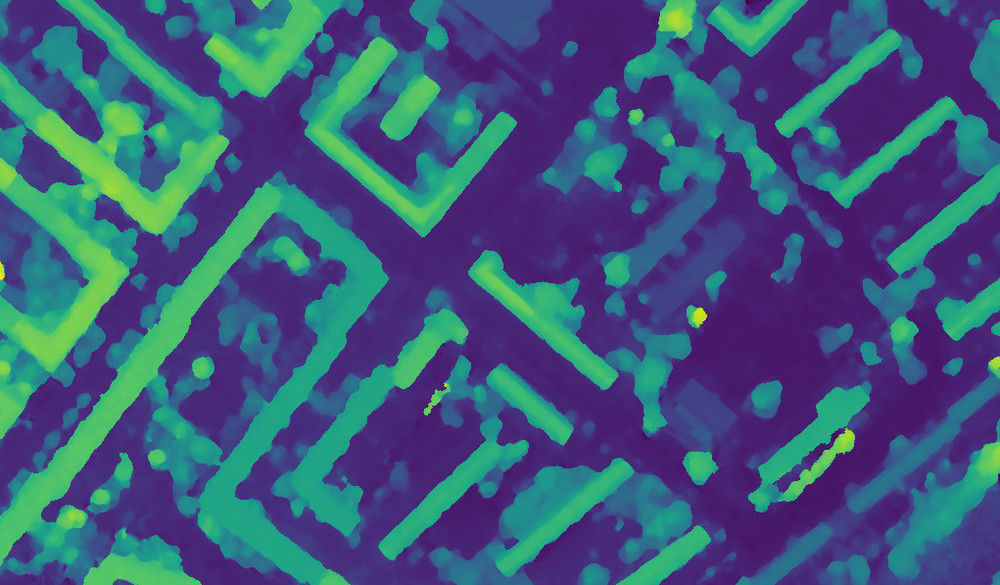}}}
    \caption{Input DSM}
  \end{subfigure}
  \hfill
  \begin{subfigure}{.23\linewidth}
    \fbox{\rotatebox{90}{\includegraphics[height=\linewidth, trim=30 0 120 0, clip]{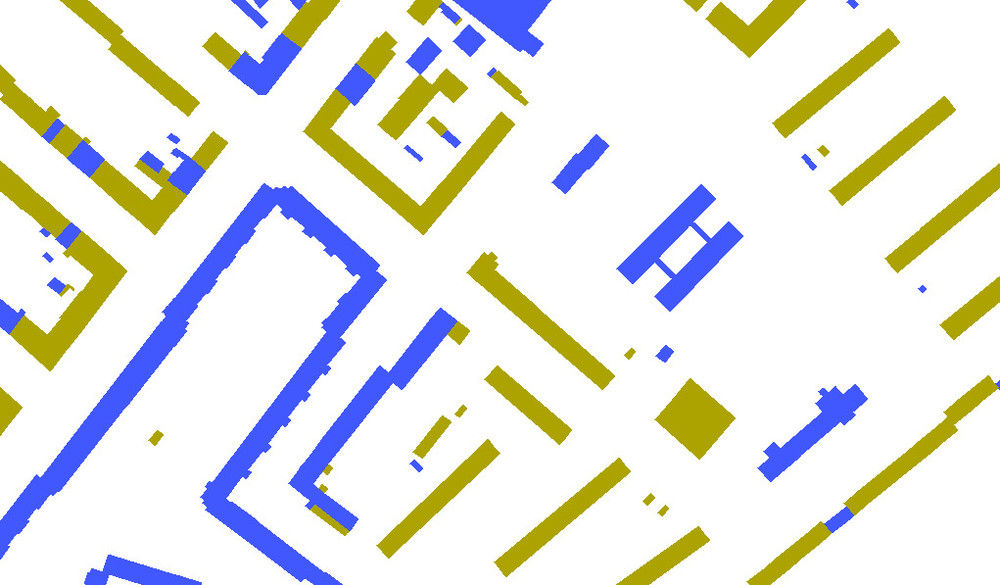}}}
    \caption{Ground-Truth}
  \end{subfigure}
  \hfill
  \begin{subfigure}{.23\linewidth}
    \fbox{\rotatebox{90}{\includegraphics[height=\linewidth, trim=30 0 120 0, clip]{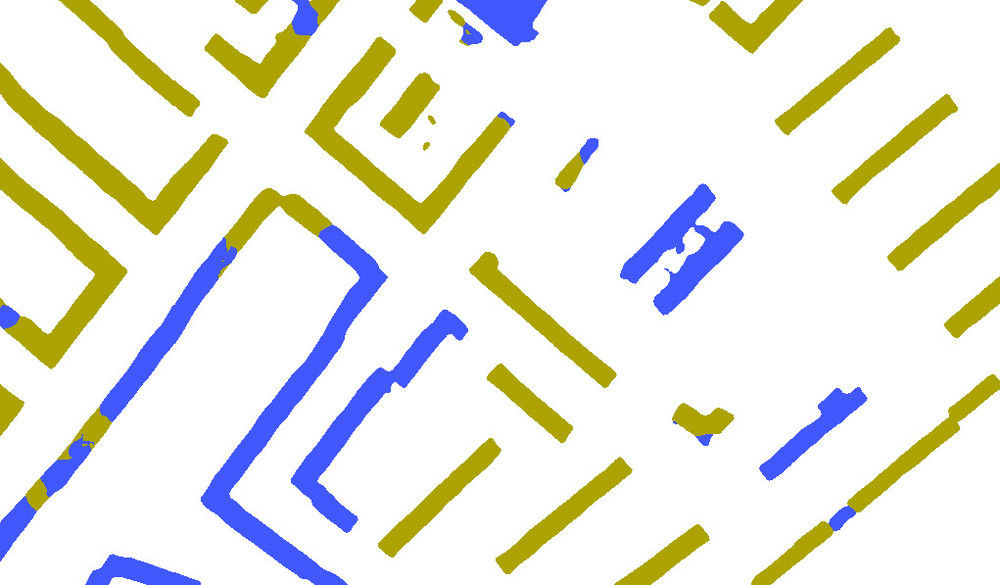}}}
    \caption{\citet{bittner2019multi}}
  \end{subfigure}
  \hfill
  \begin{subfigure}{.23\linewidth}
    \fbox{\rotatebox{90}{\includegraphics[height=\linewidth, trim=30 0 120 0, clip]{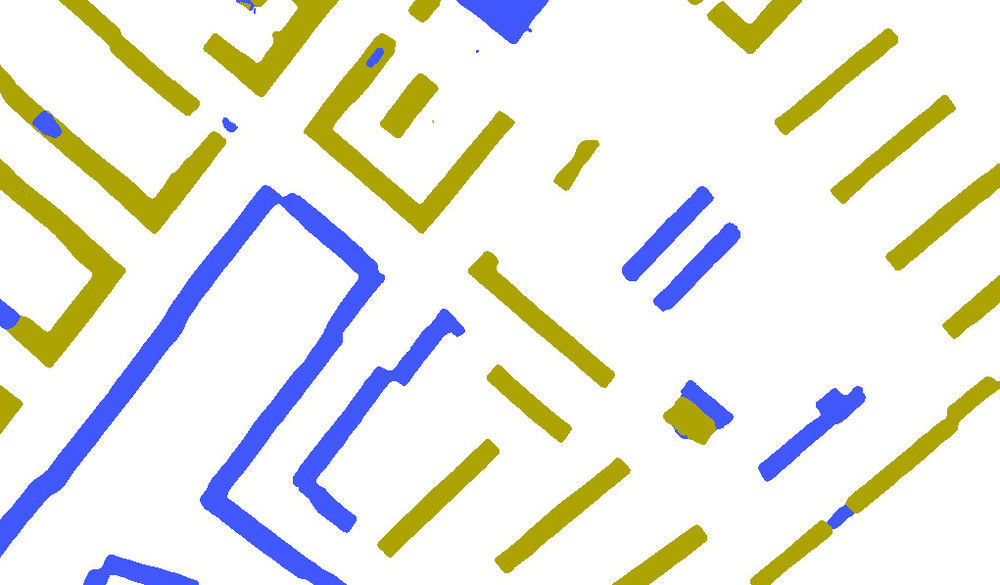}}}
    \caption{Ours}
  \end{subfigure}
  \caption{Roof type segmentation mask (b) for a scene in the Berlin test area (a) with prediction results using our approach (d) in comparison to the state of the art (c).
  Both methods succeed in segmenting building shapes.
  Classification of roof types is slightly better in our results, which is confirmed by an increase of roughly \SI{2}{\percent} in terms of mIOU to a total of \SI{65.85}{\percent}.}
  \label{fig:roof_rypes_berlin}
  \vspace{-3mm}
\end{figure}

Since all five of the tested instances of our approach achieved similar results with a difference in \gls{rmse} of only \SI{0.16}{\meter}, and particular strengths and weaknesses in different scenarios, we merged the results using a simple ensemble approach to further boost the prediction quality.
We applied equally weighted averaging over all models to produce an ensemble prediction that scores an \gls{rmse} as low as \SI{1.94}{\meter}, showing that fusing the prediction results yields a product that outperforms each of the contributing models by a sizeable margin of \SI{0.07}{\meter} to \SI{0.23}{\meter}.
Roof type masks were fused by per-pixel majority voting, resulting in a final segmentation result that yields an \gls{miou} of \SI{65.85}{\percent}, which is an improvement over the state of the art with an \gls{miou} of \SI{64.02}{\percent}.
A qualitative example of the roof type masks is shown in \cref{fig:roof_rypes_berlin}, illustrating that the achieved segmentation result improves on the state of the art.
As the ensemble further advances the state of the art in the relevant metrics (\gls{rmse} and \gls{mae}), we only considered the fused prediction results for further analyses.

\begin{figure}
    \setlength{\fboxsep}{0pt}%
    \setlength{\fboxrule}{.5pt}%
    \scriptsize

    N \raisebox{-.5mm}{\tikz{\draw[-{Stealth[scale=2.0]}] (0,0) -- (-0.01mm,0);}}
    \hfill
    Height error (clipped): \SI{-15}{\meter} \includegraphics[width=.1\linewidth, height=1ex]{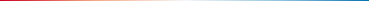} \SI{15}{\meter}

    \vspace{1mm}

    \begin{subfigure}{\linewidth}
        \fbox{%
            \rotatebox{90}{%
                \includegraphics[height=\linewidth]{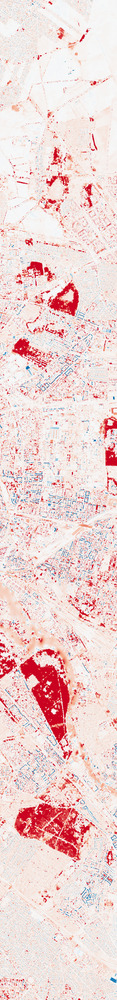}%
            }%
        }%
        \caption{Baseline}
        \label{subfig:baseline_errors}
    \end{subfigure}

    \vspace{.25cm}

    \begin{subfigure}{\linewidth}
        \fbox{%
            \rotatebox{90}{%
                \includegraphics[height=\linewidth]{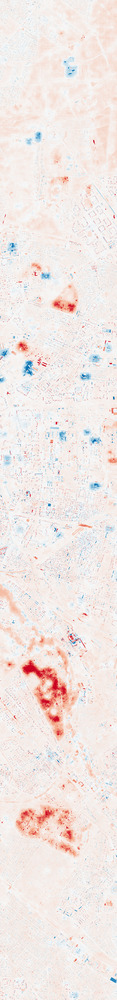}%
            }%
        }
        \caption{\Citet{bittner2018dsm}}
    \end{subfigure}

    \vspace{.25cm}

    \begin{subfigure}{\linewidth}
        \fbox{%
            \rotatebox{90}{%
                \includegraphics[height=\linewidth]{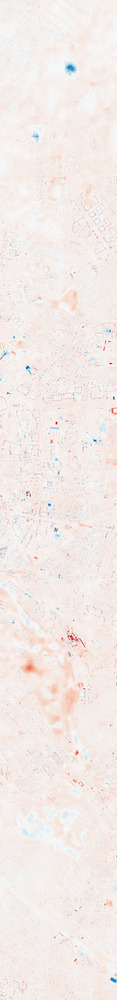}%
            }%
        }
        \caption{Ours (Ensemble)}
    \end{subfigure}

    \vspace{.25cm}

    \begin{subfigure}{\linewidth}
        \fbox{%
            \rotatebox{90}{%
                \includegraphics[height=\linewidth]{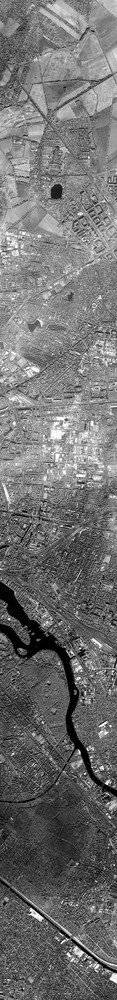}%
            }%
        }
        \caption{Optical (Pan)}
    \end{subfigure}

    \vspace{.25cm}

    \begin{subfigure}{.32\linewidth}
        \fbox{%
            \rotatebox{90}{%
                \includegraphics[height=\linewidth]{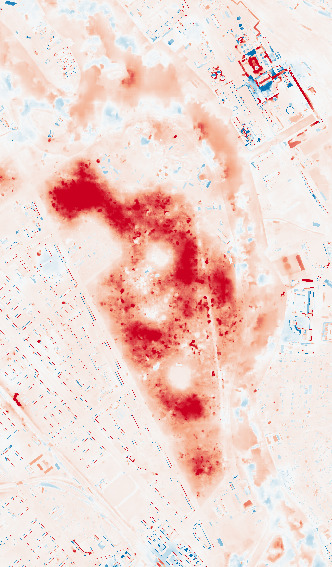}%
            }%
        }
        \caption{\Citet{bittner2018dsm}}
        \label{subfig:berlin_detail_ksenia}
    \end{subfigure}
    \hfill
    \begin{subfigure}{.32\linewidth}
        \fbox{%
            \rotatebox{90}{%
                \includegraphics[height=\linewidth]{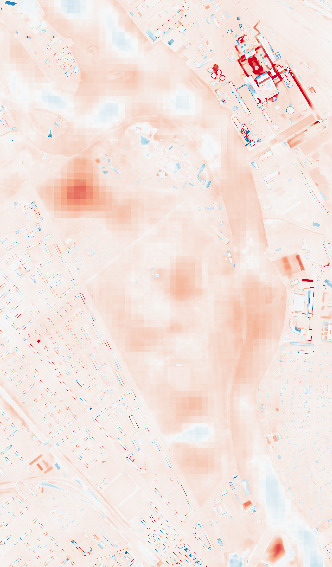}%
            }%
        }
        \caption{Ours (Ensemble)}
        \label{subfig:berlin_detail_ours}
    \end{subfigure}
    \hfill
    \begin{subfigure}{.32\linewidth}
        \fbox{%
            \rotatebox{90}{%
                \includegraphics[height=\linewidth]{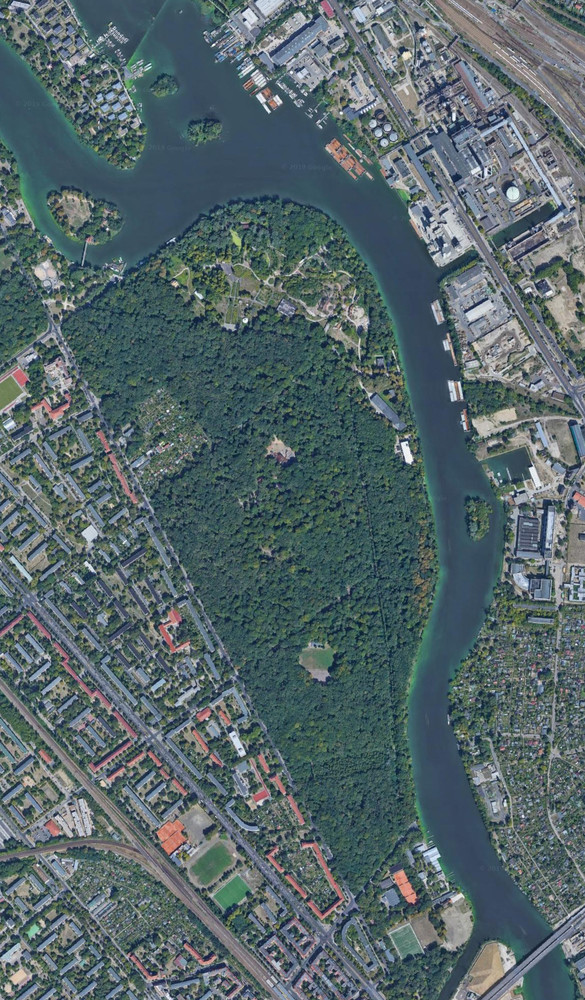}%
            }%
        }
        \caption{Optical (RGB)}
        \label{subfig:berlin_detail_rgb}
    \end{subfigure}

    \vspace{-.75cm}\raggedleft \color{white} \tiny \copyright~2019 Google~
    \vspace{.5cm}

    \caption{Remaining error in the filtered height estimates for the full Berlin test area from a baseline method (a) state-of-the-art results (b, e) and ours (c, f).
    Clearly visible are large spots with blatant errors, especially noticeable in (a) and (b).
    Our method produces remarkably better results in the affected areas, even though in some of the larger areas errors of much smaller magnitude are still evident.
    Flat and densely vegetated areas are most badly affected, as clearly visible from a detailed view (e-g).}
    \label{fig:errors_berlin}
    \vspace{-3mm}
\end{figure}

Comparing the spatial distribution of remaining errors in state-of-the-art predictions to our results, shown in \cref{fig:errors_berlin}, reveals that a fair share of the total error stems from distinct areas in the study area.
We investigated particularly notable regions in the southern part of the study area.
Qualitative evaluation exposed problems in densely vegetated areas, as shown in a detailed view in \cref{subfig:berlin_detail_rgb,subfig:berlin_detail_ksenia,subfig:berlin_detail_ours}.
Incompletely removed vegetation apparently accounts for sizeable portions of the remaining errors.
The baseline method with parameters tuned to capture trees in between blocks of buildings, as present in most of the training area, fails to remove larger patches of trees almost completely, as clearly visible from \cref{subfig:baseline_errors} already.
While this influence is also notable in both, state-of-the-art predictions and our results, our method is able to produce much more accurate height estimates, even in unfavorable scenarios.

\begin{figure*}
    \setlength{\fboxsep}{0pt}
    \setlength{\fboxrule}{.5pt}

    \begin{subfigure}{.45\linewidth}
        \includegraphics[width=\linewidth, trim=54 104 54 91, clip]{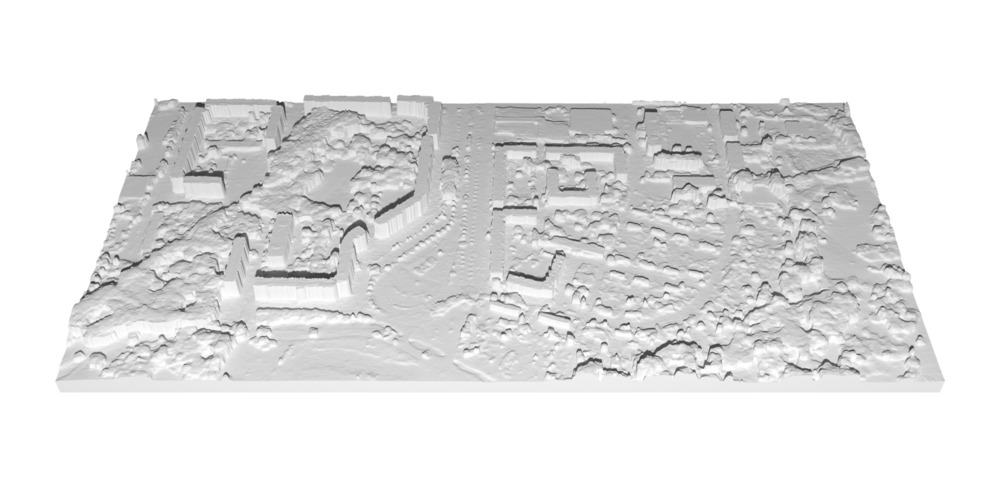} \\
        \centering \scriptsize RMSE: \SI{9.12}{\meter}
        \caption{Stereo DSM}
    \end{subfigure}
    \hfill
    \begin{subfigure}{.45\linewidth}
        \includegraphics[width=\linewidth, trim=54 104 54 91, clip]{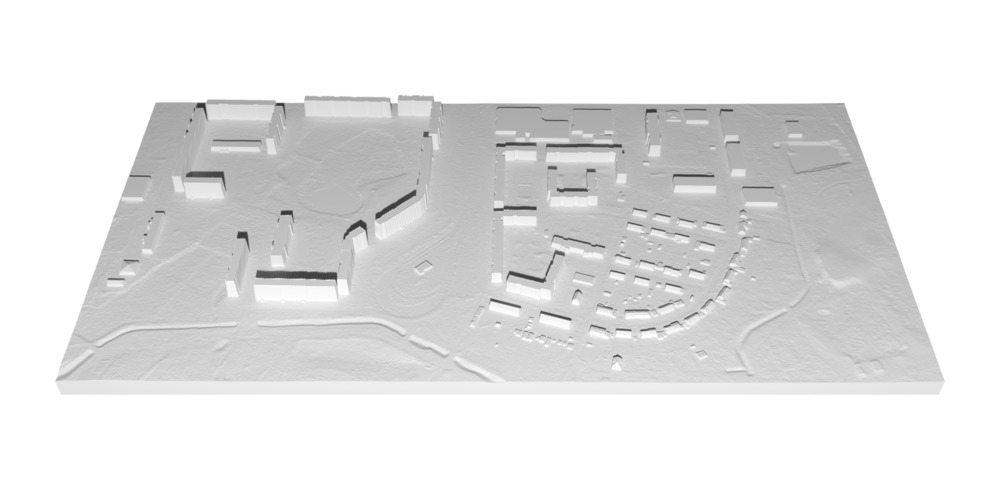} \\
        \scriptsize ~
        \caption{Ground-Truth}
    \end{subfigure}

    \vspace{.5cm}

    \begin{subfigure}{.45\linewidth}
        \includegraphics[width=\linewidth, trim=54 104 54 91, clip]{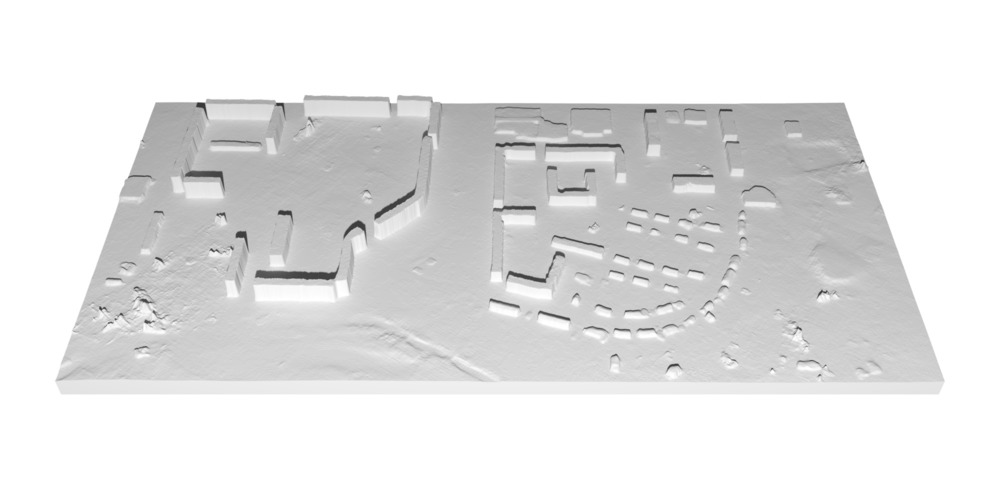} \\
        \centering \scriptsize RMSE: \SI{2.74}{\meter}
        \caption{\Citet{bittner2018dsm}}
    \end{subfigure}
    \hfill
    \begin{subfigure}{.45\linewidth}
        \includegraphics[width=\linewidth, trim=54 104 54 91, clip]{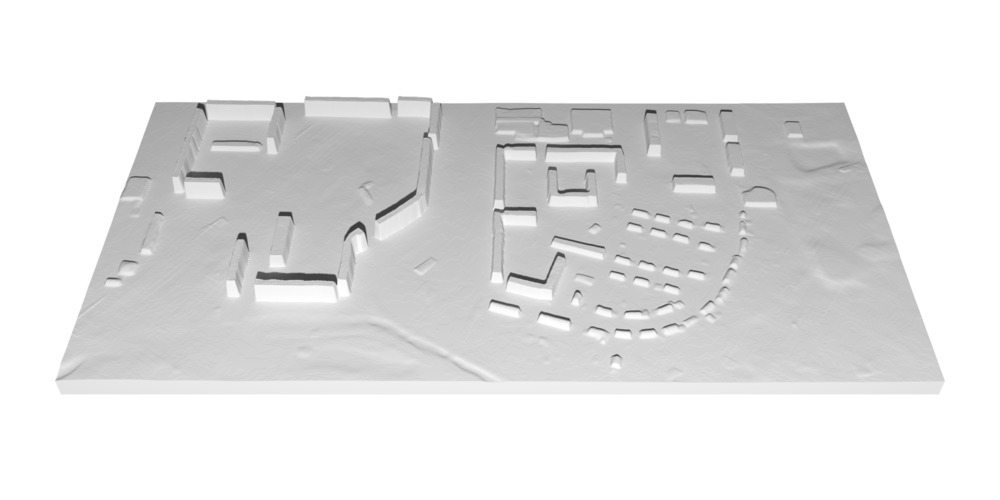} \\
        \centering \scriptsize RMSE: \SI{2.14}{\meter}
        \caption{Ours (Ensemble)}
    \end{subfigure}

    \caption{3D view of a scene depicting a building complex in Berlin-Malchow that, in addition to distinct high-rise and low-rise buildings, contains flat undeveloped land and dense vegetation.
    Dense vegetation, as clearly visible in the lower part of the scene in the stereo DSM (a), was not completely removed by a state-of-the-art method (c).
    Furthermore, some of the small buildings, see left part of the scene in the ground-truth data (b), are heavily distorted or missing.
    Our method (d) successfully removes vegetation and preserves small buildings.}
    \label{fig:test_scene_malchow}
\end{figure*}

Since the test area is under influence of a temporal distance in between the acquisition of the stereo \gls{dsm} that serves as input and the city model that provides the respective ground-truth (\cf \cref{sec:data}) a portion of the remaining errors must be accredited to this factor.
While this is acceptable for a relative comparison in large-scale areas, qualitative comparison of more detailed scenes requires addressing this effect.
We selected a small scene from the test set that contains vegetation, high rise buildings, smaller buildings, as well as undeveloped flat land and densely vegetated areas that remained unchanged in between the acquisition dates.
An analysis of such scene allows to draw qualitative conclusions and a comparison to the state of the art.
\Cref{fig:test_scene_malchow} shows a building complex in Berlin-Malchow that satisfies the desired properties, color-coded and projected into 3D space to allow for a more intuitive inspection.
From this test scene, it can be observed that both methods manage to extract and refine the relevant information from the stereo \gls{dsm} that contains a large offset in height to the ground-truth with an \gls{rmse} of \SI{9.12}{\meter}, due to noise and vegetation.
The state-of-the-art improved this baseline result by an unsurprisingly large margin to a remaining \gls{rmse} of \SI{2.74}{\meter}.
Outliers in the produced height map remain, in particular, in the densely vegetated area left of the high rise buildings, below the top-most horizontal buildings, and right of the arc of smaller buildings.
Furthermore, some of the smaller buildings, especially in the outermost line of buildings in the arc are heavily distorted and almost missing.
Our method improves on this result further by successfully suppressing the remaining outliers introduced by incomplete removal of vegetation while preserving even small buildings that are hardly visible in the stereo \gls{dsm} with the naked eye.
As a result, our predictions match the ground-truth closely and consequently achieve an even lower \gls{rmse} of only \SI{2.14}{\meter}.

\subsection{Generalization Performance}
\label{subsec:results_munich}

Evaluation of our models on the Berlin dataset allowed for a direct comparison to the state of the art.
While the test area was unseen during training and validation, it was still derived from the same stereo \gls{dsm} and is, hence, not completely independent from the training set.
We, therefore, newly introduce a second dataset, completely unseen during training, processed from different data sources, and spatially disconnected from the other study area.
The test results obtained from the Munich dataset, thus, allow for evaluating how well the developed method generalizes to unseen areas---similar to a real use-case.

\begin{figure}
  \setlength{\fboxsep}{0pt}%
  \setlength{\fboxrule}{.5pt}%
  \begin{subfigure}[t]{.3\linewidth}
      \fbox{\includegraphics[width=\linewidth]{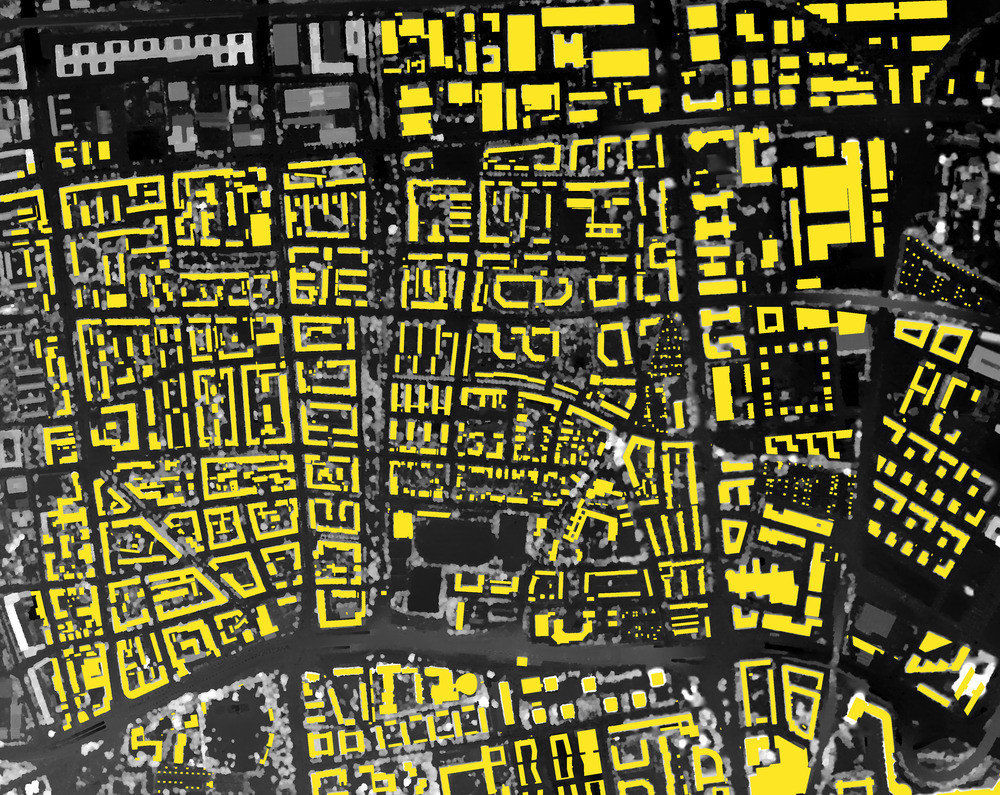}}
      \caption{Stereo DSM with Roof Mask}
  \end{subfigure}
  \hfill
  \begin{subfigure}[t]{.3\linewidth}
      \fbox{\includegraphics[width=\linewidth]{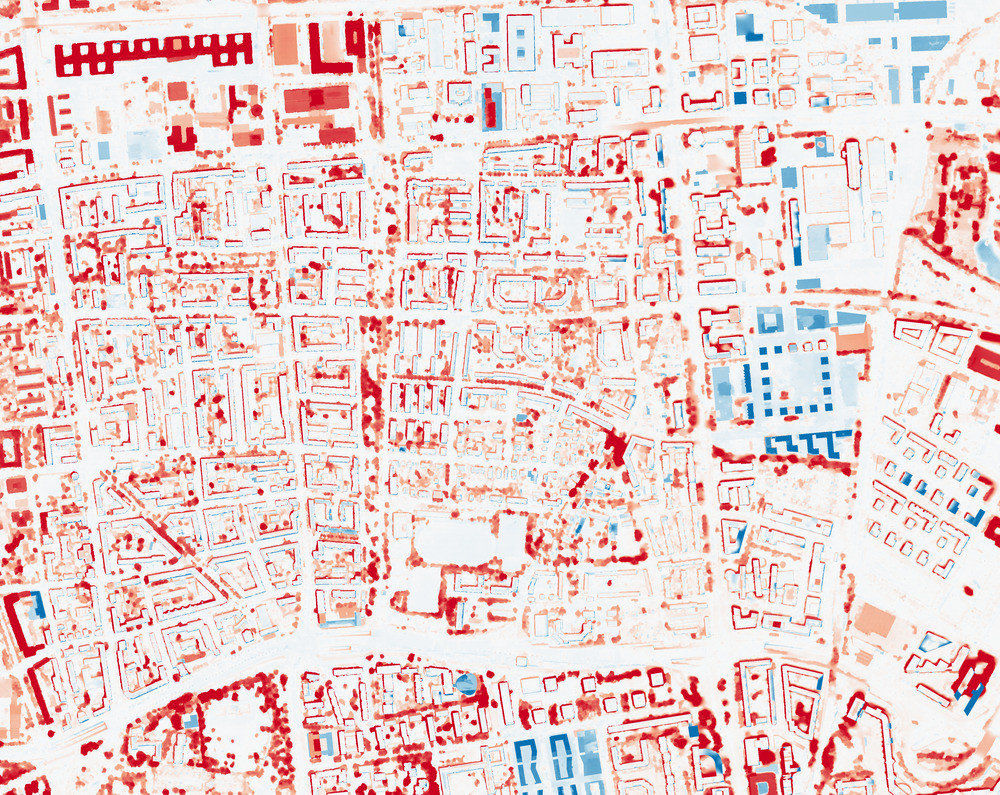}}
      \caption{Errors Stereo DSM}
  \end{subfigure}
  \hfill
  \begin{subfigure}[t]{.3\linewidth}
      \fbox{\includegraphics[width=\linewidth]{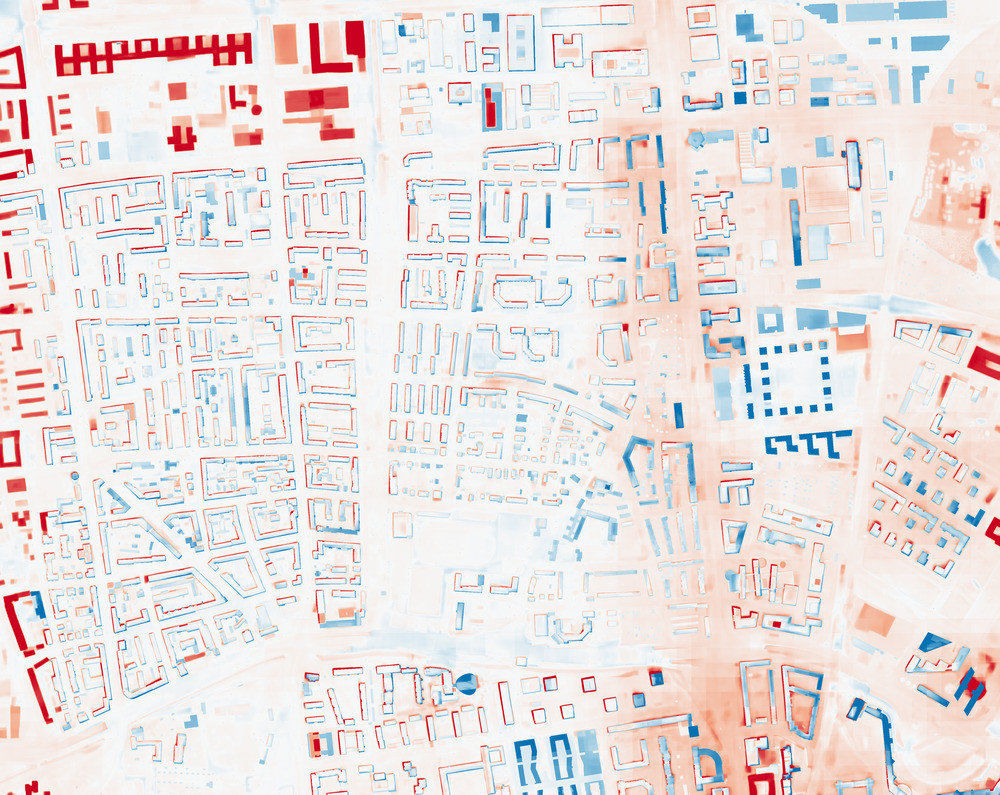}}
      \caption{Errors Prediction}
  \end{subfigure}
  \caption[]{Remaining height error in the Munich test set for stereo DSM (b) and prediction results (c).
  Comparing the error maps to a building mask \tikz{\fill[yellow] (0,0) rectangle (1.5ex,1.5ex);} and stereo DSM (a) reveals the distinct influence of newly built and demolished buildings, especially in the northwestern and eastern part of the study area.}
  \label{fig:errors_munich}
  \vspace{-3mm}
\end{figure}

\begin{figure}
    \begin{subfigure}[t]{.24\linewidth}
        \includegraphics[width=\linewidth]{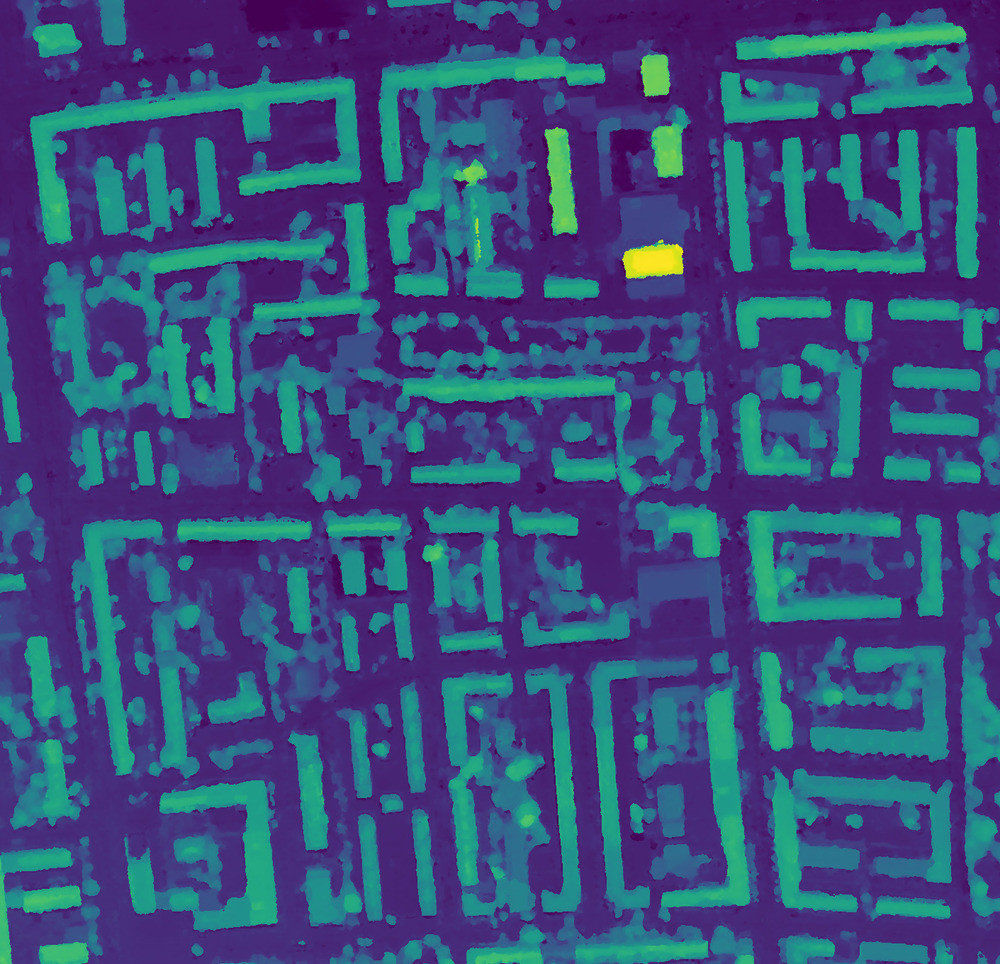} \\
        \centering \scriptsize RMSE: \SI{5.19}{\meter}
        \caption{Stereo DSM}
    \end{subfigure}
    \hfill
    \begin{subfigure}[t]{.24\linewidth}
        \includegraphics[width=\linewidth]{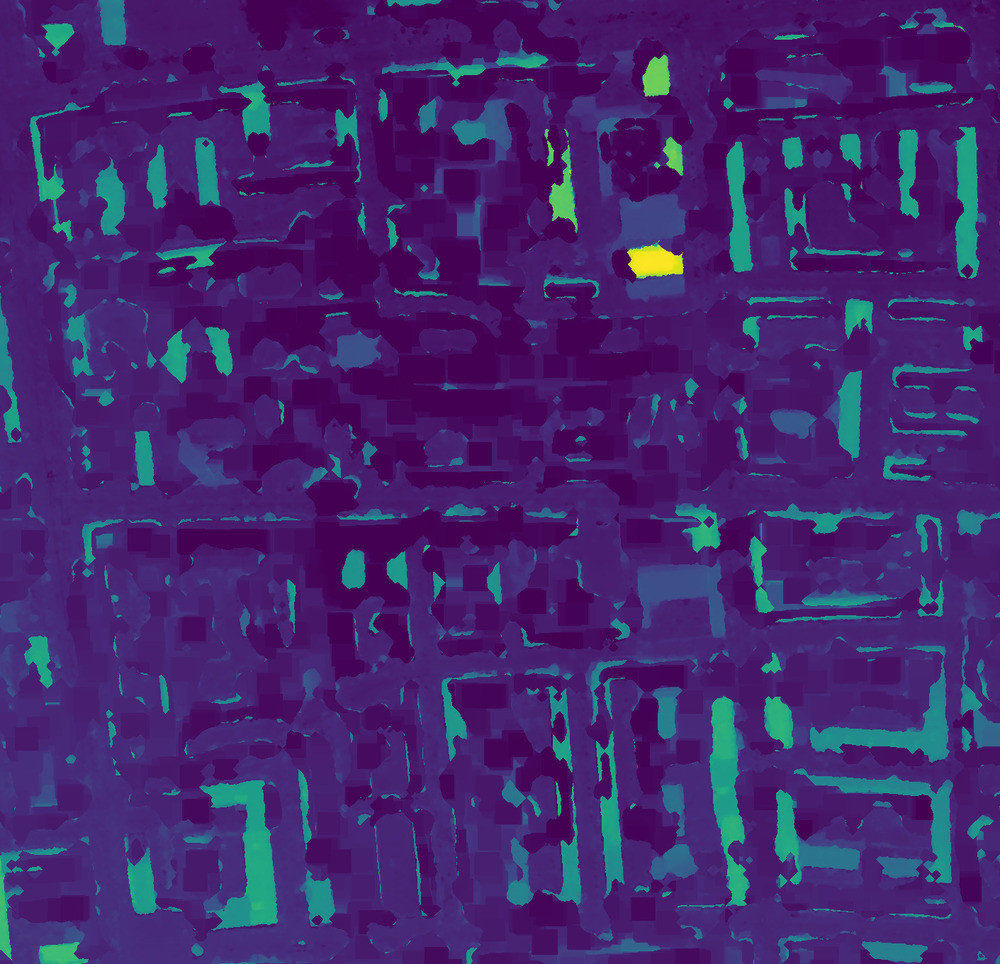} \\
        \centering \scriptsize RMSE: \SI{7.16}{\meter}
        \caption{Baseline}
    \end{subfigure}
    \hfill
    \begin{subfigure}[t]{.24\linewidth}
        \includegraphics[width=\linewidth]{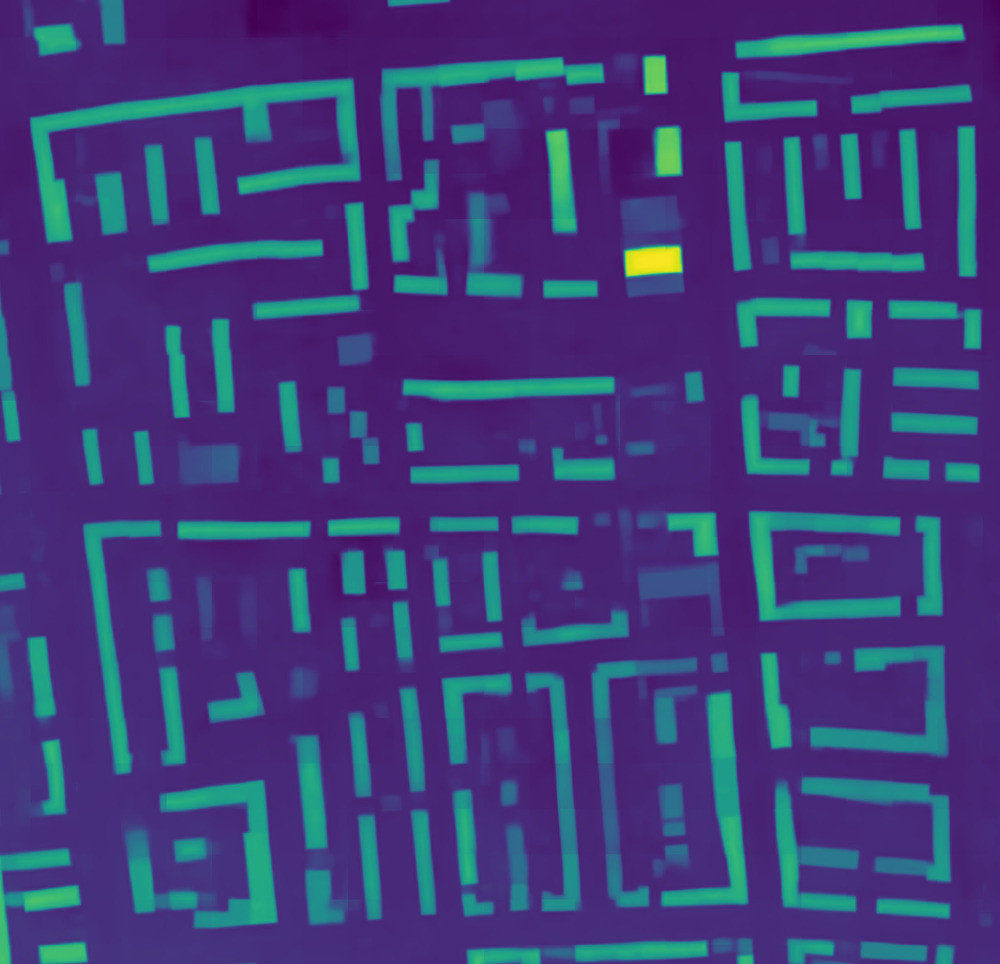} \\
        \centering \scriptsize RMSE: \SI{3.33}{\meter}
        \caption{Prediction}
    \end{subfigure}
    \hfill
    \begin{subfigure}[t]{.24\linewidth}
        \includegraphics[width=\linewidth]{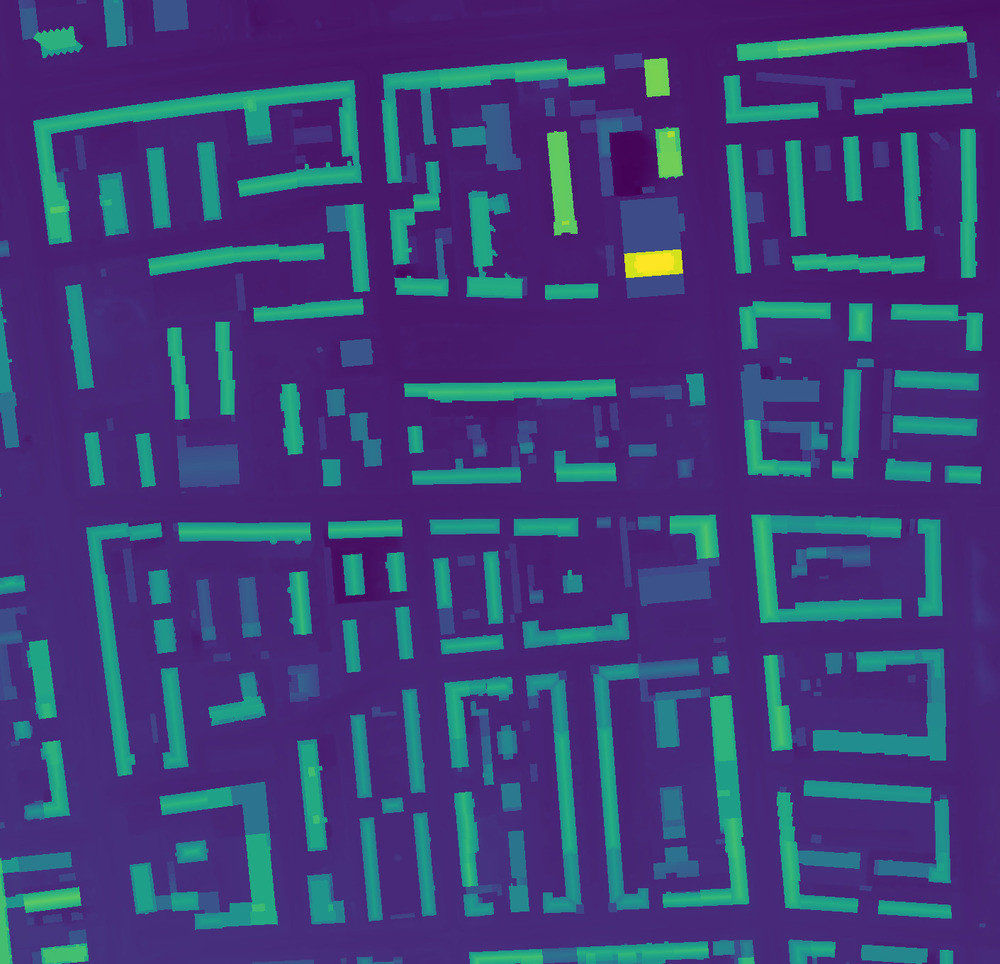} \\
        \scriptsize ~
        \caption{Ground-Truth}
    \end{subfigure}
    \caption{Predictions for an unseen area in Munich-Milbertshofen with no changes in between the acquisition of the stereo DSM and city model.
    The proposed method estimates height maps with comparable accuracy as the state of the art on the Berlin dataset, which is much closer to the training data.
    This experiment shows that networks following our concepts generalize well to unseen areas.
    The parameters of the baseline method based on hand-crafted geometric features, however, completely fail to distinguish buildings from vegetation when transferred to the unseen area.}
    \label{fig:test_scene_milbertshofen}
    \vspace{-2mm}
\end{figure}

First prediction results of the full study area yielded a surprisingly high \gls{rmse} of almost \SI{5}{\meter}, which hints at a systematic error.
Indeed, the remaining errors from both prediction and input \gls{dsm}, reveal large errors matching the shape of buildings, as to be seen in \cref{fig:errors_munich}.
As already observed and mentioned in \cref{sec:data}, a major part of the buildings in the study area was constructed, changed, or demolished in between the acquisition of input and ground-truth data.
In order to still allow for a meaningful quantitative analysis, we extracted the central block of buildings from the study area, that remained unchanged.
From this scene in Munich-Milbertshofen, shown in \cref{fig:test_scene_milbertshofen}, we derived similar performance metrics as before.
With an \gls{rmse} of \SI{3.33}{\meter}, the overall performance is slightly worse than the results on the Berlin dataset.
Yet, it still easily outperforms the very simple stereo \gls{dsm} baseline by almost \SI{2}{\meter}.
Interestingly, the baseline method which is based on hand-crafted features and thresholding parameters tuned on the training set of the Berlin study area, fails to correctly distinguish vegetation from buildings here and therefore scores an error of \SI{7.16}{\meter}.
Thus, it even degrades the quality of the original \gls{dsm}.
A qualitative evaluation shows that our method successfully extracts and refines buildings and removes vegetation.
Even in this completely independent study area, located in an absolute height of \SI{519}{\meter} above sea level, which is roughly \SI{500}{\meter} higher than Berlin, with a \gls{dsm} processed from images acquired by a different sensor, our method still performs well.
With an \gls{rmse} that is still on par with state-of-the-art results on the much less independent Berlin test area, the proposed method generalizes well to unseen data, as oposed to the more traditional baseline method.

While this independent testing protocol allows drawing first conclusions about the generalization performance, further analysis is desirable.
The considered cities of Berlin and Munich share certain properties with regard to building types and appearance.
CityGML models, utilized as ground-truth in our experiments, are not available globally rendering evaluation on a larger scale infeasible.
However, our evaluation on independent test data approves that there are no signs of overfitting to the training set.

In future work, \gls{lidar}-derived data could potentially serve as ground-truth for predicted height maps.
Building footprints could be evaluated using reference data from topographic maps, such as the OpenStreetMap project.
Adopting such data sources for the evaluation process, however, requires further considerations regarding data pre-processing and the validation routine, and are therefore considered out of scope here.

\section{Discussion}

The experimental evaluation of the proposed approach, presented in \cref{sec:evaluation}, showed the overall convincing performance of instances of our method.
The general suitability of multi-task multi-objective training, as already known from prior work, was confirmed for our generalized modular encoder-decoder approach with learned weighting terms in an ablation study, presented in \cref{sec:experiments}.
Here, adding supervision through additional objectives and tasks consistently improved validation performance.
This is especially noteworthy, as the additional ground-truth data was automatically derived from the same source.
Hence, it is not adding to the effort of acquiring a suitable dataset consisting of various compatible labels for a single source image, which is a major obstacle to multi-task learning in general.

In a study with different instances of our model, DeepLabv3+ decoders for the main task achieved the best performance overall, despite featuring the lowest capacity out of all considered decoder architectures.
While UNet decoders can achieve competitive results, they are more demanding in terms of memory as they contain a higher number of parameters.
The PSPNet decoder, in comparison, was not able to produce satisfactory results for \gls{dsm} filtering.
It does, however, yield best results for the roof type segmentation task.
Even though both of the other decoders are on par with this performance, the PSPNet decoder has the advantage of using few parameters and no skip connections from the encoder.
Considering this, the DeepLabv3+/PSPNet model that performed best on the test set is most versatile and could easily be deployed in combination with other encoders that may not support multiple skip connections.
Counterintuitively, the performance of the tested models was not directly related to their capacity.

The learned weighting terms successfully balanced the different objectives, taking away the need for tedious manual tuning.
The results of our studies suggest adding even more objectives and tasks for future approaches.
Manually tuning the weighting terms to find an optimal set would surely render this strategy infeasible quickly, whereas an automatic and dynamic weighting procedure scales to this challenge easily.

\section{Conclusions}
\label{sec:conclusion}

We presented a generalized framework for filtering stereo \glspl{dsm} of urban areas, obtained from satellite imagery, via modular multi-task encoder-decoder \glspl{cnn}.
Our models utilize multiple objectives evaluating geometrical properties, an adversarial term, and segmentation results of a secondary roof type segmentation task.
Balancing of these objectives in the multi-task multi-objective function, subject to optimization, was done by learned task uncertainty-based weights.
Prior approaches on this application using multi-task learning \citep{bittner2019multi} can be modeled as instances of our framework.

In extensive experiments, we studied the performance of a variety of specific instances from the class of models our approach defines.
Especially, different architectures for the decoder modules and the influence of the employed objectives were subject to studies.
Our models were trained and tested on a common dataset from Berlin to allow for a fair comparison to the state of the art and were further evaluated on completely unseen and independent test data from a different city to give insight to their generalization performance as in a real use-case.

The proposed method consistently outperforms the simple stereo \gls{dsm} baseline, the stronger baseline based on hand-crafted geometric features, and the current state of the art by a sizeable margin of more than \SI{1}{\meter} RMSE, which corresponds to a decrease of around \SI{34}{\percent}.
In ablation studies, we were able to show the influence of each objective and the weighting scheme on the final result, and the suitability of different decoder architectures.
Overall, the best performing instance of our model utilizes the full set of investigated objective functions for the main \gls{dsm} filtering task, namely an \lone{} loss, a surface normal loss, a conditional adversarial loss term, and a segmentation objective for the secondary roof type classification task.
The single objectives were combined into multi-task multi-objective loss function using learned balancing terms based on homoscedastic uncertainty.
Decoder modules based on DeepLabv3+ and PSPNet were employed for the main and auxiliary tasks.
In comparison to a basic setup that was trained using a simple \lone{} regression loss only, the proposed model not only yields a much lower error but also requires roughly \SI{40}{\percent} less training, despite containing a higher number of parameters due to the addition of a secondary decoder.

Source code and configuration files for the experiments reported in this paper will be publicly available online soon\footnote{\url{https://github.com/lukasliebel/multitask_dsm_filtering}}.


\section*{Acknowledgements}
The work of Lukas Liebel was funded by the German Federal Ministry of Transport and Digital Infrastructure (BMVI) under reference 16AVF2019A.


\printbibliography

\end{document}